\documentclass[runningheads]{llncs}

\usepackage{eccv}

\usepackage{eccvabbrv}

\usepackage{graphicx}
\usepackage{booktabs}
\usepackage{makecell}
\usepackage{multirow} 
\usepackage{wrapfig}

\usepackage[accsupp]{axessibility}  

\usepackage{hyperref}

\usepackage{orcidlink}

\begin{document}

\title{FreeMotion: MoCap-Free Human Motion Synthesis with Multimodal Large Language Models} 

\titlerunning{FreeMotion}

\author{Zhikai Zhang\inst{1,3} \and
Yitang Li\inst{1,3} \and
Haofeng Huang\inst{1} \and 
Mingxian Lin \inst{3} \and 
Li Yi \inst{1,2,3}}

\authorrunning{Z.~Zhang et al.}

\institute{$^1$~Tsinghua University~~ $^2$~Shanghai AI Laboratory~
$^3$~Shanghai Qi Zhi Institute}

\maketitle

\begin{figure}
  \includegraphics[width=\linewidth]{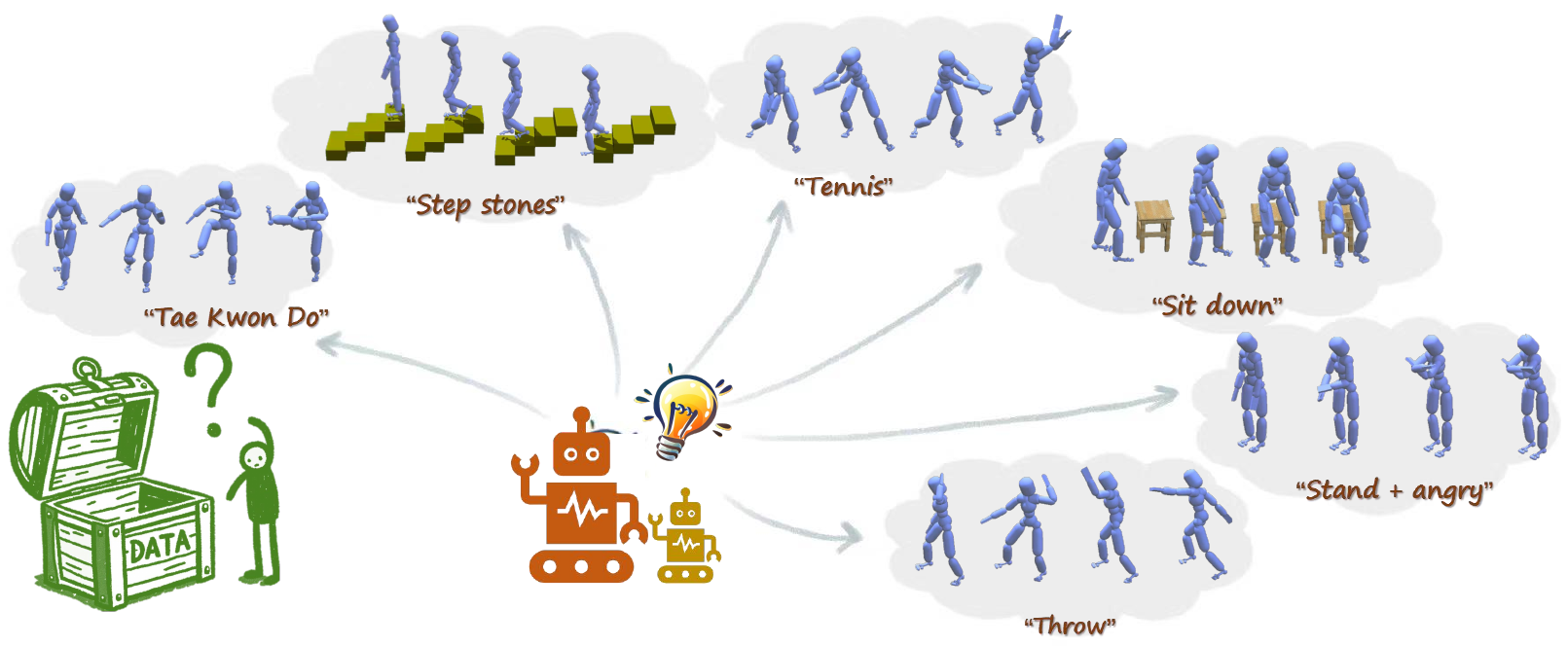}
  \caption{Our method for the first time, without any motion data, explores open-set human motion synthesis using natural language instructions as user control signals based on MLLMs across any motion task and environment.  }
  \label{fig:teaser}
\end{figure}

\begin{abstract}
Human motion synthesis is a fundamental task in computer animation. Despite recent progress in this field utilizing deep learning and motion capture data, existing methods are always limited to specific motion categories, environments, and styles. This poor generalizability can be partially attributed to the difficulty and expense of collecting large-scale and high-quality motion data. At the same time, foundation models trained with internet-scale image and text data have demonstrated surprising world knowledge and reasoning ability for various downstream tasks. Utilizing these foundation models may help with human motion synthesis, which some recent works have superficially explored. However, these methods didn't fully unveil the foundation models' potential for this task and only support several simple actions and environments. In this paper, we for the first time, \textit{without any motion data}, explore open-set human motion synthesis using natural language instructions as user control signals based on MLLMs across any motion task and environment. Our framework can be split into two stages: 1) sequential keyframe generation by utilizing MLLMs as a keyframe designer and animator; 2) motion filling between keyframes through interpolation and motion tracking. Our method can achieve general human motion synthesis for many downstream tasks. The promising results demonstrate the worth of mocap-free human motion synthesis aided by MLLMs and pave the way for future research. 
\end{abstract}

\section{Introduction}
\label{sec:intro}

Synthesizing humanoid movements and interactions is a cornerstone for advancing embodied AI, enhancing the realism of video games, enriching experiences in VR/AR, and empowering robots with the ability to interact with humans. Therefore, researchers have been long seeking automatic ways for humanoid animation synthesis.
Existing works~\cite{tevet2022human, guo2022generating,hassan2021stochastic,jiang2023motiongpt,tevet2022motionclip,xiao2023unified,yao2022controlvae,hassan2023synthesizing,li2023object,zhang2022motiondiffuse,zhang2023motiongpt} have made significant progress facilitated by reference motion trajectories depicting real human movements collected through motion capture (mocap) systems. While these methods have yielded high-fidelity animations, they are intrinsically limited by the scope of the mocap data. Due to the inherent difficulty and expenses of motion capturing, the largest publicly accessible mocap datasets~\cite{mahmood2019amass, guo2022generating} only encompass dozens of hours of motion, which is still far from enough to cover the vast array of daily human motions. As such, data-driven animation synthesis is usually confined to pre-recorded motion datasets and lacks open-set generalizability to novel environments and unseen human behaviors. 

In the realm of machine learning, Multimodal Large Language Models (MLLMs) have recently emerged as a transformative force, showcasing remarkable competency in inferring and adapting to open-set scenarios. These models have been successful across a spectrum of tasks that range from perception~\cite{2023GPT4VisionSC, yang2023dawn, dong2023dreamllm} and high-level planning~\cite{huang2023voxposer, hu2023look} to low-level manipulative actions~\cite{yu2023language, ma2023eureka}. This success prompts us to consider whether we can leverage powerful MLLMs trained on internet-scale image and text data (e.g., GPT-4V~\cite{2023GPT4VisionSC}) to break free from the dependency on mocap data and instead generate open-set humanoid animations that can dynamically adapt to new and ever-changing environments and tasks.

In this work, we for the first time demonstrate MLLMs’ ability for open-set humanoid motion synthesis controlled by natural language user input \textit{without any motion data}. (See Fig.~\ref{fig:teaser}.)
Directly applying MLLMs trained on image and text data as motion generators is not proper, as they may not capture the subtleties of continuous motion necessary for realistic character animation. Nonetheless, MLLMs excel in understanding high-level action narratives and keyframes, akin to a lead animator's role in traditional studios. We, therefore, propose to first leverage MLLMs to decompose open-set humanoid motion into narrative plots and corresponding keyframes and then in the second stage develop automatic motion filling algorithms to close the gap between the discrete understanding of MLLMs and the continuous nature of humanoid movement. 

Specifically, in the first stage, we employ two specialized GPT-4V agents to generate a sequence of keyframes. One agent acts as the keyframe designer, using text descriptions of the desired motion (e.g., walking) and the current state of the humanoid (e.g., the left leg advancing while the right leg is stationary), along with a rendered image of the humanoid, to predict the text for the subsequent keyframe (e.g., the left leg making contact with the ground as the right leg begins to lift) and the time interval between two adjacent frames. The other agent, the keyframe animator, is then presented with this predicted text. With a set of pre-defined commands to manipulate the humanoid's joints and the current state information, the animator selects appropriate commands to adjust the humanoid's pose to match the designer's description, using the rendered images for visual feedback. This pose adjustment may be refined multiple times. The designer and animator collaborate in this iterative fashion until they complete the motion sequence.

In the second stage, to transform a sequence of keyframes into a fluid motion clip, we engage in motion filling. Initially, we perform interpolation on the keyframe sequence to create a time-continuous motion clip. However, since interpolation may not adhere to physical laws, we utilize a motion tracking policy that corrects for physically implausible poses and transitions. Drawing inspiration from successful model-based tracking methods~\cite{yao2022controlvae, fussell2021supertrack,won2022physics}, we develop a CVAE-based policy empowered by an MLP-based world model to track the interpolated motion. Unlike previous approaches confined to flat terrain, we integrate height maps to inform our policy and world model about varying terrain, ensuring our synthesis is adaptable to diverse environments.

We evaluate our method on a wide range of downstream tasks, including motion synthesis, style transfer, human-scene interaction, and stepping stones. Our method achieves surprising results \textit{without any motion data}.

\vspace{-3mm}
\section{Related Work}
\vspace{-3mm}
\subsection{Foundation Models for Motion Synthesis}
\vspace{-3mm}
Striking advancements of foundation models~\cite{radford2021learning, chowdhery2023palm, bommasani2021opportunities, 2023GPT4VisionSC, brown2020language,dong2023dreamllm,ouyang2022training, team2023gemini} have been made during the past few years. The success of LLMs stimulated interest in MLLMs (Multimodal Large Language Models)~\cite{2023GPT4VisionSC, team2023gemini}, which extends LLMs to accept visual input. They either learn from visual signal and text 
simultaneously from scratch~\cite{team2023gemini} or employ a cross-modal connector to align the features of visual encoders to the LLM's text embedding space. As foundation models demonstrate impressive capabilities on many downstream applications these years, researchers try to ground their knowledge into motion synthesis tasks. Generating reward functions for particular tasks is adopted by several works~\cite{yu2023language, ma2023eureka, ma2023liv} as an intermediate interface connecting motion instructions and physics-based motion controllers. Methods based on reward design utilize GPT's ability of logical reasoning and code generation. However, only a small set of motions is suitable to be represented as a reward function. These methods fail to maintain their performance when applied to open-set motion synthesis. Rather than designing task-specific reward functions, ~\cite{rocamonde2023vision} utilizes CLIP~\cite{radford2021learning} to compute the similarity between observation and motion text, which serves as the reward value for policy training. It only supports the simplest human motions (e.g. sitting, raising hands). Compared with existing methods, our method takes the first step toward open-set motion synthesis based on MLLMs. 
\vspace{-3mm}
\subsection{Human Motion Synthesis}
\vspace{-2mm}

Human motion synthesis is a fundamental task in computer animation. With the popularity of neural networks and motion capture data, data-driven methods have become mainstream~\cite{tevet2022human, guo2022generating,hassan2021stochastic,jiang2023motiongpt,tevet2022motionclip,xiao2023unified,yao2022controlvae,hassan2023synthesizing,li2023object,zhang2022motiondiffuse,zhang2023motiongpt, zhao2023synthesizing, zhao2022compositional, xu2023interdiff, wang2022humanise, wei2023enhanced}. Recent researchers use generative models to recover kinematic motions from Gaussian noise, given various conditional signals. VAE-based methods~\cite{holden2016deep, petrovich2022temos,guo2022generating,ghosh2023imos} and GAN-based~\cite{li2022ganimator, liu2021aggregated, hassan2023synthesizing} methods have been widely explored during the past few years.
\cite{holden2016deep} employs a Variational Autoencoder (VAE) to acquire a general motion manifold, enabling the synthesis and editing of character motion based on high-level control parameters. 
Also, employing diffusion models in motion synthesis \cite{tevet2022human,zhang2022motiondiffuse,yuan2023physdiff,rempe2023trace, xu2023interdiff, wei2023enhanced} has emerged as a new trend due to their state-of-the-art performance. 
Several recent studies have explored novel approaches in motion control and synthesis. ~\cite{peng2022ase} distills a large set of expert policies into a latent space for a high-level controller. ~\cite{won2022physics} employs a conditional VAE for expert demonstration mimicry, while ~\cite{yao2022controlvae} uses a CVAE for flexible skill representation and policy learning. Inspired by GAIL, ~\cite{peng2021amp} develops a discriminator to ensure style consistency and task-specific reward but still compliance in motion data.
\vspace{-2mm}
\section{Method}
\vspace{-1mm}
\begin{figure}[t]
 	\includegraphics[width=\linewidth]{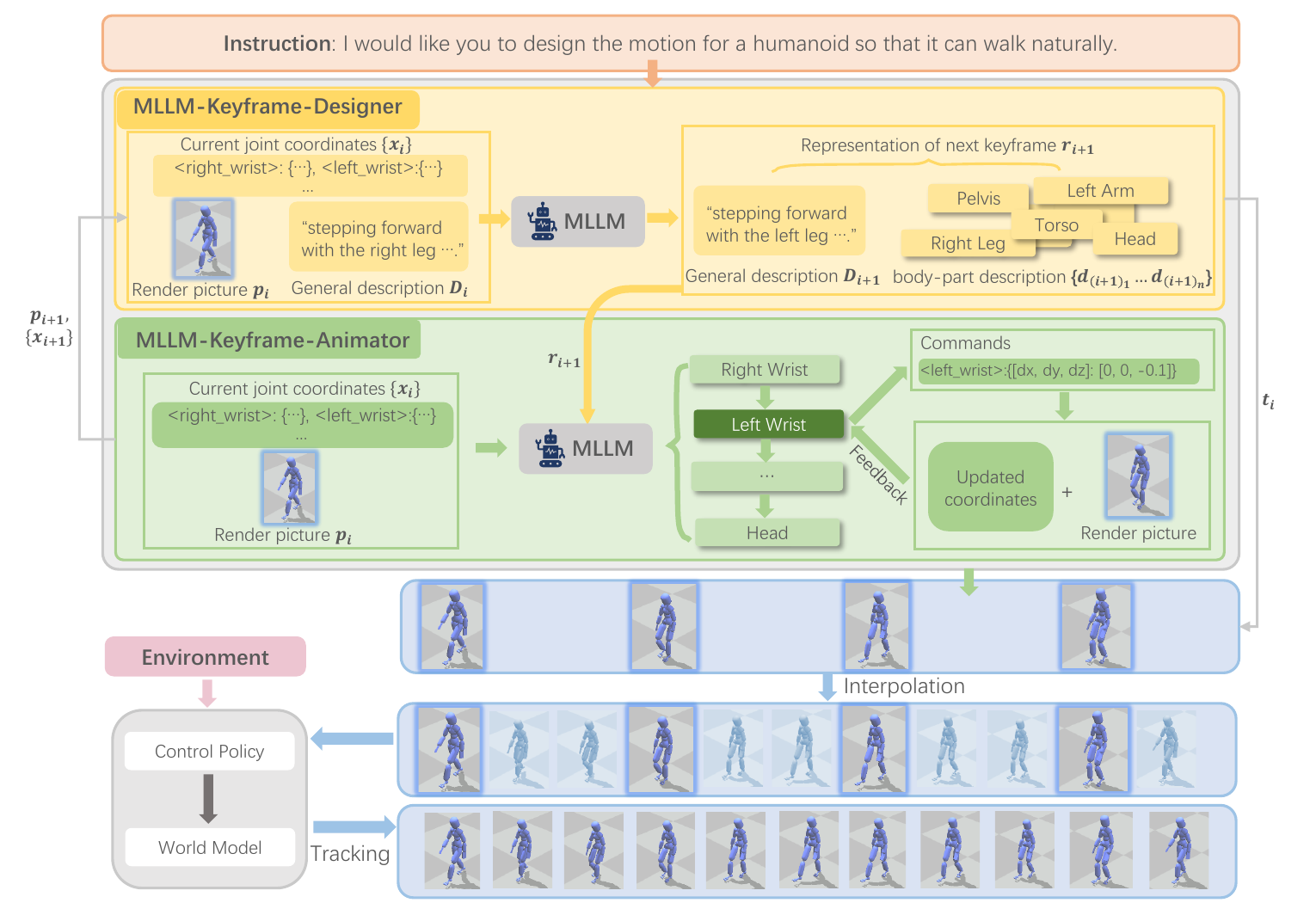}	
	\caption{\textbf{Overview of FreeMotion.} FreeMotion adopts two specialized GPT-4V agents for sequential keyframe generation. Then we utilize interpolation and environment-aware motion tracking to fill the blank between keyframes.
	}
	\label{fig:overview}
\vspace{-5mm}
\end{figure}
One straightforward way to generate human motion clips from MLLMs is to utilize MLLMs as motion state predictors. However, this method often yields unsatisfactory results due to the underrepresentation of such motion data in the MLLMs’ training corpora and the subtleties of continuous motion. MLLMs excel in their world knowledge and logical reasoning ability drawn from internet-scale text and image data. However, these abilities only come into play in high-level semantic space rather than low-level motion space. A significant challenge is bridging this gap and effectively applying the MLLMs' capabilities to motion space. To solve this problem, we propose our framework, FreeMotion, and split the problem into two stages: 1) sequential keyframe generation by utilizing MLLMs as a designer and animator; 2) motion filling between keyframes through interpolation and motion tracking. The underlying insight of our method is the utilization of MLLMs solely within the high-level
semantic space. Since keyframes in a motion usually contain richer and more salient semantic information, we utilize MLLMs to decompose a given motion temporally and spatially by generating sequential keyframes. The blank between keyframes is left for motion-filling techniques, including interpolation and environment-aware motion tracking. The overview of our method is shown in Fig.~\ref{fig:overview}.
\vspace{-3mm}
\subsection{Sequential Keyframe Generation from MLLMs}
\vspace{-2mm}
Given a user instruction requiring a specific motion, we hope MLLMs can translate it into a sequence of humanoid poses, each representing a keyframe in the motion. Such a task involves motion understanding, keyframe reasoning, and humanoid posture adjustment. As it is challenging for the MLLM to output a correct sequence simultaneously, we employ two specialized GPT-4V agents, each playing a distinct role. One serves as a keyframe designer, aiming to translate the input motion instruction into relatively low-level body part descriptions of sequential keyframes. The other one acts as a keyframe animator, who takes the description of one keyframe generated by the designer and fits a humanoid's pose to the description through visual feedback using a set of pre-defined pose adjustment commands. We will discuss the details of each GPT-4V agent in the following sections. 
\vspace{-5mm}
\subsubsection{Keyframe Designer.} The keyframe designer’s role is to translate high-level motion instruction $\boldsymbol{I}$ into a sequence of more detailed, low-level keyframe representation $\boldsymbol{R} = \{\boldsymbol{r_{1}},\ldots,\boldsymbol{r_{m}}\}$, where $m$ is the number of keyframes to represent the motion. Each $\boldsymbol{r_{i}}$ is composed of a general full-body description $\boldsymbol{D_{i}}$ (e.g., \textit{"The humanoid is stepping forward with its left leg, the right leg is stationary and the arms are swinging opposite to the legs."}) and a series of body-part (e.g., left arm, right leg, torso) descriptions $\{\boldsymbol{d_{i_{1}}},\ldots,\boldsymbol{d_{i_{n}}}\}$ (e.g., \textit{"The left arm is moving backwards in a smooth arc, with the shoulder back, the elbow slightly bent, and the hand relaxed."}), where $n$ is the number of body parts. Each time, the keyframe designer depicts the next keyframe given the full-body description $\boldsymbol{D_{i}}$, a rendered picture $\boldsymbol{p_{i}}$ of the current humanoid, the humanoid's current joint coordinates $\{\boldsymbol{x_{i}}\}$, and the motion instruction $\boldsymbol{I}$ as input, outputting: 1) the low-level keyframe representation of the next keyframe $\boldsymbol{r_{i+1}}$; 2) the time interval $\boldsymbol{t_{i}}$ (e.g., \textit{"0.5s"}) between the current state and the predicted next keyframe. Starting with $\boldsymbol{D_{0}}$ (\textit{"The humanoid is standing on the ground."}), the keyframe designer can produce the whole sequence of keyframe representation $\boldsymbol{R}$. This process spatially and temporally decomposes the high-level motion instruction $\boldsymbol{I}$, integrating the MLLM’s knowledge into a more tangible motion representation for the subsequent keyframe animator. The MLLM can make a reasonable motion-to-keyframe decomposition without the rendered picture $\boldsymbol{p_{i}}$ of the humanoid. But the presentation of $\boldsymbol{p_{i}}$ offers the keyframe designer a chance to better understand the humanoid's current state and make an improved representation $\boldsymbol{r_{i+1}}$ of the next keyframe.  

At this stage, the MLLM plays a crucial role in determining the spacing between adjacent keyframes. Excessively distant keyframes can lead to unstable results and result in motion artifacts even with physics correction. Conversely, overly close keyframes can make the generation process excessively tedious, diminishing the keyframes' ability to provide constructive guidance. However, in most of our experiments, GPT-4V successfully generates feasible keyframes. This success could be attributed to the MLLM's inherent understanding of motion dynamics; it comprehends that a motion sequence is comprised of several distinct stages. For instance, in walking, the sequence involves lifting the left foot, stepping forward, setting down the left foot, and lifting the right foot. Given this understanding, the MLLM is able to generate an appropriate number of keyframes, effectively segmenting the entire motion sequence.

We let the keyframe designer to automatically determine the termination point of motion design. It is instructed to signal the completion of the entire motion sequence by outputting "Done" once it perceives the completion of the specified non-periodic motion, or believes that a periodic motion has concluded after a full cycle. Besides leveraging the MLLM's capability to recognize motion termination, we also manually set an upper limit to ensure the motion design won't be endless.
\vspace{-5mm}
\subsubsection{Keyframe Animator.} Provided with a detailed next-keyframe representation $\boldsymbol{r_{i+1}}$, joint coordinates $\{\boldsymbol{x_{i}}\}$, and the rendered picture $\boldsymbol{p_{i}}$, the keyframe animator is responsible for adjusting a humanoid's pose $\boldsymbol{s^k_i}$ to fit the representation $\boldsymbol{r_{i+1}}$. Adjustments are made in order of body parts listed by the keyframe designer. Rather than directly tuning the joint's position or rotation, we regularize the adjustment as a set of commands, each corresponding to a specific joint movement. These commands are implemented using kinematic methods, such as forward and inverse kinematics. This regularization not only frees MLLM from the tedium of tuning spatial features joint by joint but also gives semantic information to pose adjustments so that the reasoning ability can be utilized. All commands used by GPT-4V are listed in Tab.~\ref{tab:command_set}. We also allow the GPT-4V animator to rotate the camera around the humanoid to observe body parts of interest better.

Despite the simplifications that have been made, it remains a complex task for the MLLM to accurately adjust the pose in a single attempt. Luckily, visual signals can serve as feedback and make multi-iteration adjustment possible. Concretely, given a certain body part and the representation $\boldsymbol{r_{i+1}}$, the MLLM chooses one of the commands and outputs corresponding parameters to adjust the body part. When the command is executed on the humanoid, the updated joint coordinates and the rendered picture are passed to the keyframe animator as feedback. This loop continues until the animator believes the body part's pose aligns with the descriptions or the times of this body part's adjustment meet the upper limit, which is 5 in our method. When the adjustment is finished for one body part, the animator switches to the next body part according to a pre-defined order. The animator finally passes the updated coordinates $\{\boldsymbol{x_{i+1}}\}$ and an updated rendered picture $\boldsymbol{p_{i+1}}$ for the next keyframe back to the designer when the adjustment for every body part is completed. 

It is worthing to note that although our method incorporates a visual feedback mechanism, it typically converge within the upper limit for a single body part. This is primarily because most body parts either remain static or experience only minor alterations during transitions. Consequently, the total number of adjustments required by the animator to transition the humanoid from $\boldsymbol{s^k_{i}}$ to $\boldsymbol{s^k_{i+1}}$ consistently remains under 10.
\begin{table}
\vspace{-5mm}
\caption{\textbf{Command Set.} We regularize the pose adjustment as a set of commands.}
\vspace{-6mm}
\label{tab:command_set}
\begin{center}
\resizebox{0.9\textwidth}{!}{
\begin{tabular}{ll}
  \toprule
    Command & Function\\ \midrule
  Single joint movement  & \makecell[l]{move a selected joint around its parent joint to a target place}  \\ \midrule
  End effector movement  &  \makecell[l]{move a selected end effector quickly to a target place through \\pre-defined IK chains}\\ \midrule
  \makecell[l]{Pelvis rotation/movement with \\support points on the ground} & \makecell[l]{rotate/move the pelvis with one or more support points on the \\ground through IK}\\ \midrule
  \makecell[l]{Pelvis rotation/movement without \\support points on the ground} & \makecell[l]{rotate/move the pelvis without support points on the ground \\ through direct rotation/movement}\\ \midrule
  Single joint roll  &  roll a selected joint\\ \midrule
  Camera rotation & rotate the camera around the humanoid\\
  \bottomrule
\end{tabular}}
\end{center}
\vspace{-10mm}
\end{table}
\vspace{-2mm}
\subsection{Motion Filling through Interpolation and Motion Tracking}
\vspace{-3mm}
\begin{figure*}
\vspace{-6mm}
	\centering	
 	\includegraphics[width=0.87\linewidth]{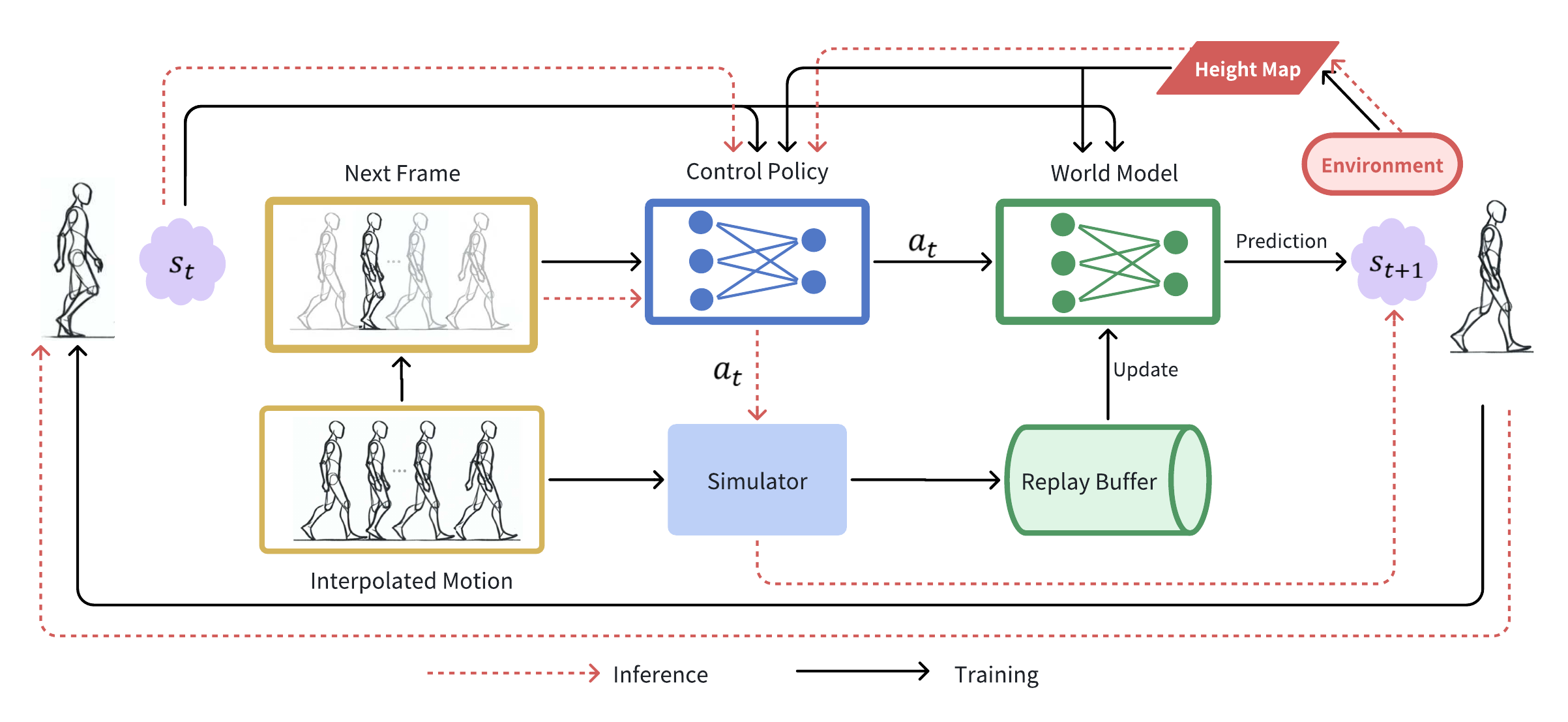}	
        \vspace{-2mm}
	\caption{\textbf{Policy training and inference.} We incorporate height maps as visual signals, enabling our policy and world model to be aware of diverse environmental conditions.
	}
	\label{fig:policy}
\vspace{-6mm}
\end{figure*}
After obtaining a series of keyframes with each keyframe specified by humanoid poses $\{\boldsymbol{s_{1}^k},\ldots,\boldsymbol{s_{m}^k}\}$ and time intervals $\{\boldsymbol{t_{1}},\ldots,\boldsymbol{t_{m-1}}\}$ based on our instructions, we perform linear position and rotation interpolation on these keyframes to achieve continuous motion frames, resulting in an interpolated frame rate of 20 frames per second. 
However, straightforward interpolation may fall short of ensuring the motion's physical validity. To address this, we turn to model-based motion tracking methods, as proven in ~\cite{ren2023insactor}, which successfully navigate the challenges of infeasible state transitions. We implement a refined motion tracking system using a CVAE-based policy combined with an MLP-based world model, drawing inspiration from the methodology of ControlVAE~\cite{yao2022controlvae}. A notable innovation in our study is the integration of environmental signals, enhancing the model's responsiveness to dynamic contexts. The methodology is depicted in Fig.~\ref{fig:policy}.
\vspace{-4mm}
\subsubsection{Environment Visual Signals Extraction.}

Our incorporation of height maps as visual signals enables our policy and world model to be cognizant of the environment and thus to be environment-aware. We derive a height map around the humanoid pelvis from the current environment observation and flatten it into a vector $\boldsymbol{o_{t}}$ at simulation time step $t$.
\vspace{-3mm}
\subsubsection{CVAE-based Motion Control Policy.}

Our CVAE-based motion control policy is formulated as a conditional encoder and decoder. The state $\boldsymbol{s}$ at each simulation time step can be fully characterized by ${\{\boldsymbol{x_{j}}, \boldsymbol{q_{j}}, \boldsymbol{v_{j}}, \boldsymbol{\omega_{j}}\}}, j \in B$, where $B$ is the set of rigid bodies and $\boldsymbol{x_{j}, q_{j}, v_{j}, \omega_{j}}$ stand for the position, orientation, linear velocity, and angular velocity of each rigid body, respectively. Given the current state $\boldsymbol{s_{t}}$ and the interpolated trajectory $\Tilde{\tau} = \{\boldsymbol{\Tilde{s}_{1}},\ldots,\boldsymbol{\Tilde{s}_{T}}\}$, we first encode the state transition $(\boldsymbol{s_{t}}, \boldsymbol{\Tilde{s}_{t+1}})$ and visual signals $\boldsymbol{o_{t}}$ into a latent variable $\boldsymbol{z}$.  
The network for encoding is referred to \( q_\phi \), parameterized by \( \phi \), which models the embedding to a Gaussian distribution:
\begin{equation}
q_\phi (\boldsymbol{z_t}|\boldsymbol{s_t}, \boldsymbol{\Tilde{s}_{t+1}},\boldsymbol{o_t}) = \mathcal{N} (\boldsymbol{z_t}; \mu_\phi (\boldsymbol{s_t}, \boldsymbol{\Tilde{s}_{t+1},o_t)}, \Sigma_\phi (\boldsymbol{s_t}, \boldsymbol{\Tilde{s}_{t+1}},\boldsymbol{o_t})).
\end{equation}

Using the latent variable derived from the previous network, we can generate an action by the decoder. Our decoder can be formulated as a conditional distribution $p(\boldsymbol{a}|\boldsymbol{s},\boldsymbol{z})$ that outputs an action $\boldsymbol{a}$ according to the character's current state $\boldsymbol{s}$ and a latent variable $\boldsymbol{z}$. We model the policy $p_\theta$ parameterized by \( \theta \) as a Gaussian
distribution as well:
\begin{equation}
p_\theta(\boldsymbol{a_t}|\boldsymbol{s_t},\boldsymbol{z_t})=\mathcal{N} (\boldsymbol{a_t};\mu_\theta (\boldsymbol{s_t},\boldsymbol{z_t}), \Sigma_\theta (\boldsymbol{s_t},\boldsymbol{z_t})).
\end{equation}
\vspace{-8mm}
\subsubsection{MLP-based World Model}
We approximate true transition probability distribution \( p(\boldsymbol{s_{t+1}}|\boldsymbol{s_t}, \boldsymbol{a_t}) \) in the simulator using an environment-aware world model \( \omega(\boldsymbol{s_{t+1}}|\boldsymbol{s_t}, \boldsymbol{a_t}, \boldsymbol{o_t}) \), which is another Gaussian distribution
\begin{equation}
\omega(\boldsymbol{s_{t+1}}|\boldsymbol{s_t}, \boldsymbol{a_t}, \boldsymbol{o_t}) \sim \mathcal{N}(\boldsymbol{s_{t+1}};\mu_{\omega}(\boldsymbol{s_t}, \boldsymbol{a_t}, \boldsymbol{o_t}), \Sigma_{\omega}(\boldsymbol{s_t}, \boldsymbol{a_t}, \boldsymbol{o_t})).
\end{equation}
\vspace{-10mm}
\subsubsection{Inference}
At each time step \( t \), with $\boldsymbol{s_t}, \boldsymbol{\Tilde{s}_{t+1}}, \boldsymbol{o_{t}}$ as input, our policy outputs $\boldsymbol{a_{t}}$ to the simulator for the computation of $\boldsymbol{s_{t+1}}$. Starting with $\boldsymbol{s}_0=\boldsymbol{\Tilde{s}}_0$ and continuously repeating this process, we obtain a trajectory \( \tau = \{\boldsymbol{s}_0, \boldsymbol{s}_1, \ldots, \boldsymbol{s_{T-1}}, \boldsymbol{s_T}\} \) where the character keeps moving under the guidance of the given interpolated motion frames. 
\vspace{-5mm}
\subsubsection{Training Process.}
Detailed training process and loss terms are complicated and not the focus of our work. We adopt almost the same training process and loss terms as ControlVAE~\cite{yao2022controlvae}. We recommend readers refer to the original paper for more details. It is worth noting that we do not train a motion tracker for every single generated motion since it's very time-consuming. For each downstream task, which will be presented in the next section, we concatenate all interpolated motions together to train a policy and world model. The collection of each trajectory is conducted within each interpolated motion so that it doesn't span different motions. The length of each interpolated motion must meets the minimum rollout length for successful training. Therefore motions that don't meet the requirement will be padded with its last frame.

\vspace{-5mm}
\section{Tasks}
\vspace{-3mm}
We evaluated our methods on various downstream tasks across different motion categories and environments, including motion synthesis, style transfer, human-scene interaction, and stepping stones. We use ODE~\cite{smith2005open} for physical simulation. 
\vspace{-8mm}
\subsection{Motion Synthesis}
\vspace{-2mm}
In this task, we evaluate our method's performance on motion synthesis. We conducted two experiments. In the first, we compared our method with two recent data-driven methods, MDM~\cite{tevet2022human} and MLD~\cite{chen2023executing} on HumanAct12. In the second, we compared our method to zero-shot motion synthesis methods~\cite{tevet2022motionclip, hong2022avatarclip}, using motions that were unseen by both the baselines and our model for testing.
\vspace{-5mm}
\subsubsection{Baseline}
\vspace{-2mm}
MDM~\cite{tevet2022human} and MLD~\cite{chen2023executing} are recent data-driven methods trained on HumanAct12~\cite{guo2020action2motion}, which is an action-to-motion dataset, containing 12 action categories and 1191 motion clips. It is worth noting that some actions in HumanAct12 involve interactions with objects, e.g., Drink, Lift dumbbell, Turn steering wheel. GPT-4V can imagine the existence of these virtual objects and generate corresponding keyframes. The test motions listed in Tab.~\ref{tab:normal_motion_synthesis} are seen for baseline methods and unseen for MLLMs during development.

Zero-shot motion synthesis has been explored by some CLIP-based methods~\cite{tevet2022motionclip, hong2022avatarclip}. To evaluate the ability of our method to tackle this task utilizing the world knowledge of MLLMs, we compared the performance on Olympic sports following the setting in MotionCLIP~\cite{tevet2022motionclip} to~\cite{tevet2022motionclip, hong2022avatarclip}, excluding motions not suitable for physical tracking on the ground, for example, Cycling and Diving. In this experiment, corresponding motion data is unseen for both baseline methods and ours. 

\begin{wraptable}{r}{6cm}
\vspace{-4mm}
\caption{\textbf{Motion Synthesis on HumanAct12.} FreeMotion achieves good results without motion data.}
\vspace{-4mm}
\label{tab:normal_motion_synthesis}
\begin{center}
\resizebox{0.9\linewidth}{!}{
\begin{tabular}{lccc}
  \toprule
 & \multicolumn{3}{c}{\textit{User Study}}\\
    & MDM~\cite{tevet2022human} & MLD~\cite{chen2023executing} & Ours \\ \midrule
  Warm up    & 26.00\% &\pmb{38.00\%} & 36.00\% \\
  Walk   & 10.00\% & 22.00\% & \pmb{68.00\%} \\
  Run   & 30.00\% & 32.00\% & \pmb{38.00\%} \\
  Jump   & 16.00\% & 28.00\% & \pmb{56.00\%}\\
  Drink   & 14.00\% & \pmb{46.00\%} & 40.00\%\\
  Lift\_dumbbell & 26.00\% & 32.00\% & \pmb{42.00\%}\\
  Sit  & 30.00\% & \pmb{44.00\%} & 26.00\%\\
  Eat  & 22.00\% & 30.00\% & \pmb{48.00\%}\\
  Turn\_steering\_wheel & 32.00\% & 28.00\% & \pmb{40.00\%}\\
  Phone  & 30.00\% & 32.00\% & \pmb{38.00\%}\\
  Boxing  & 16.00\% & 24.00\% & \pmb{60.00\%}\\
  Throw   & 20.00\% & 14.00\% & \pmb{66.00\%}\\ \midrule
  Average & 22.67\% & 30.83\% & \pmb{46.50\%}\\
  \bottomrule
\end{tabular}}
\vspace{-11mm}
\end{center}
\end{wraptable}
\vspace{-2mm}
\subsubsection{Metric}
\vspace{-2mm}
Given those commonly-used inception models for evaluation, as in~\cite{tevet2022human, chen2023executing}, are overfitting on their training datasets and fail to evaluate our method, we choose to utilize user preference as many prior works~\cite{tevet2022motionclip, aberman2020unpaired, hong2022avatarclip} where inception models are unavailable. We asked 50 volunteers to perform a user study in terms of two focuses: 1) the consistency with input texts, and 2) motion quality (physical feasibility, naturalness, etc.). We show the volunteers with randomly sampled motions generated by the same prompt (an action category in this experiment) from our method and baseline methods side by side. The volunteers are asked to select the one with the best performance according to the above two focuses.
\vspace{-2mm}
\subsubsection{Analysis}
\begin{figure*}
\vspace{-5mm}
	\centering	
 	\includegraphics[width=0.8\linewidth]{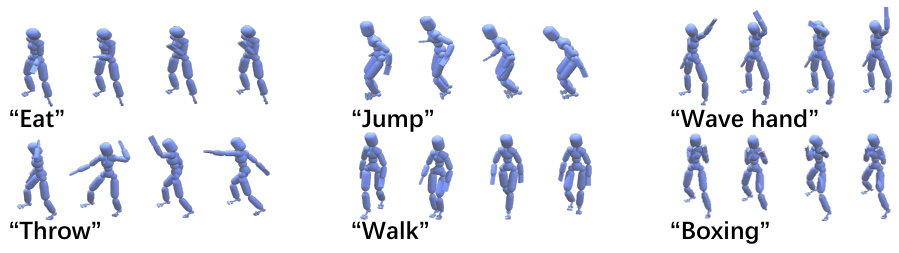}	
	\caption{\textbf{Motion synthesis visualization results of FreeMotion on HumanAct12.} FreeMotion can synthesize realistic motions across different categories.
	}
	\label{fig:humanact12}

\vspace{-5mm}
\end{figure*}

\vspace{-2mm}
The results of motion synthesis on HumanAct12 are shown in Tab.~\ref{tab:normal_motion_synthesis}. It is evident that FreeMotion outperforms traditional data-driven methods in most cases. The key factor contributing to this success is FreeMotion's ability to maintain physical plausibility, a challenge for methods like MDM and MLD. Fig.~\ref{fig:humanact12} highlights FreeMotion's capability to generate realistic, previously unseen motions on HumanAct12.

\begin{figure*}
\vspace{-3mm}
	\centering	
 	\includegraphics[width=0.9\textwidth]{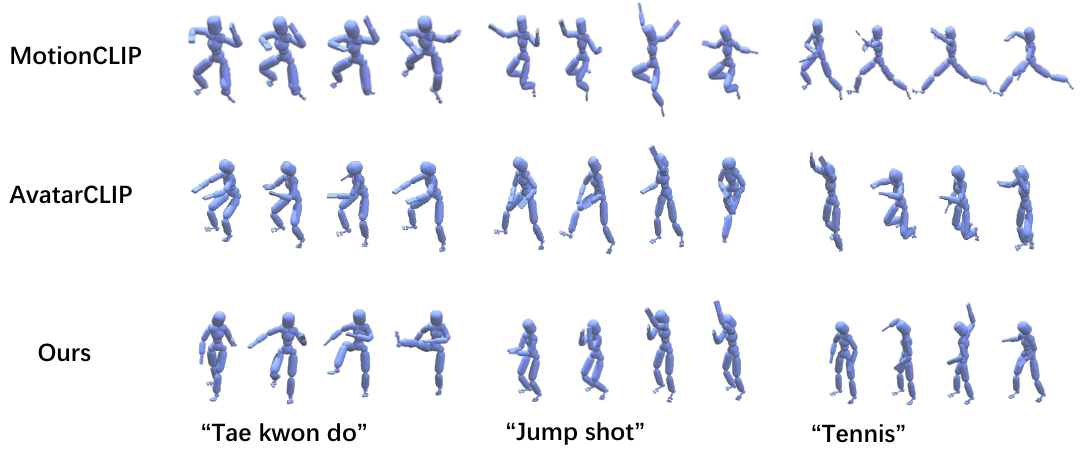}	
	\caption{\textbf{Motion synthesis visualization results on Olympic sports.} FreeMotion can synthesize satisfactory motions even on challenging Olympic sports.
	}
	\label{fig:sports}
\vspace{-3mm}
\end{figure*}

For the second experiment, the user preference score is shown in Tab.~\ref{tab:openworld_motion_synthesis}. Our method outperforms two existing zero-shot motion synthesis methods significantly. Fig.~\ref{fig:sports} also vividly illustrates that FreeMotion is capable of synthesizing realistic Olympic sports motions, whereas MotionCLIP and AvatarCLIP tend to produce unnatural motions. Despite MotionCLIP and AvatarCLIP benefiting from CLIP's zero-shot generalizability, they fall short in adhering to physical constraints and accurately interpreting the composition and sequence of motions. A case in point is a jump shot in basketball: ideally, the player first lifts the ball to the chest before jumping. However, these methods often struggle to replicate this sequential accuracy. 

\begin{table*}[]
\caption{\textbf{Olympic Sports.} FreeMotion surpasses existing methods significantly.}
\vspace{-2mm}
\label{tab:openworld_motion_synthesis}
\resizebox{\textwidth}{!}{%
\begin{tabular}{l|l|ccccccccccc}
\hline
           Metrics& Methods & \raisebox{-.5\height}{\includegraphics[width=1cm]{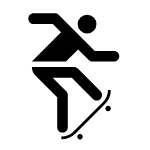}} & \raisebox{-.5\height}{\includegraphics[width=1cm]{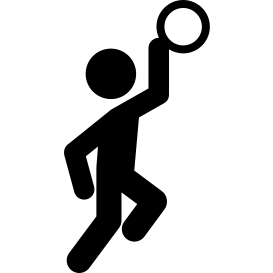}} & \raisebox{-.5\height}{\includegraphics[width=1cm]{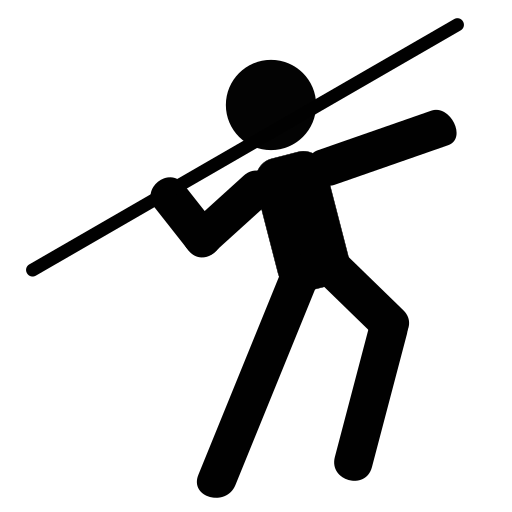}} & \raisebox{-.5\height}{\includegraphics[width=1cm]{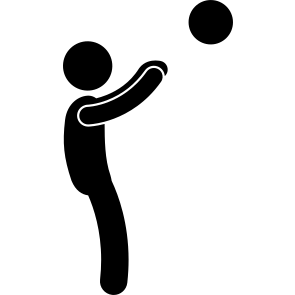}}    & \raisebox{-.5\height}{\includegraphics[width=1cm]{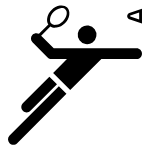}} & \raisebox{-.5\height}{\includegraphics[width=1cm]{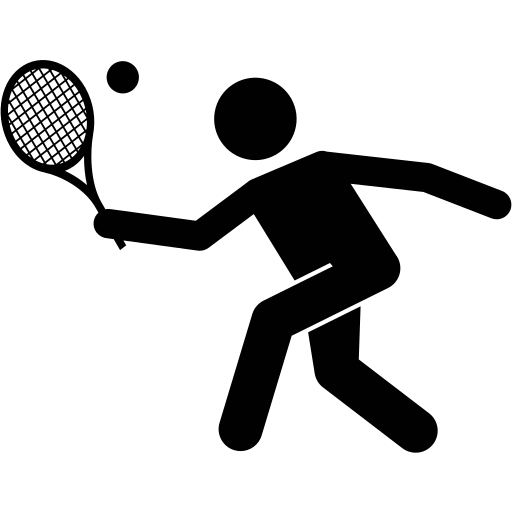}} & \raisebox{-.5\height}{\includegraphics[width=0.95cm]{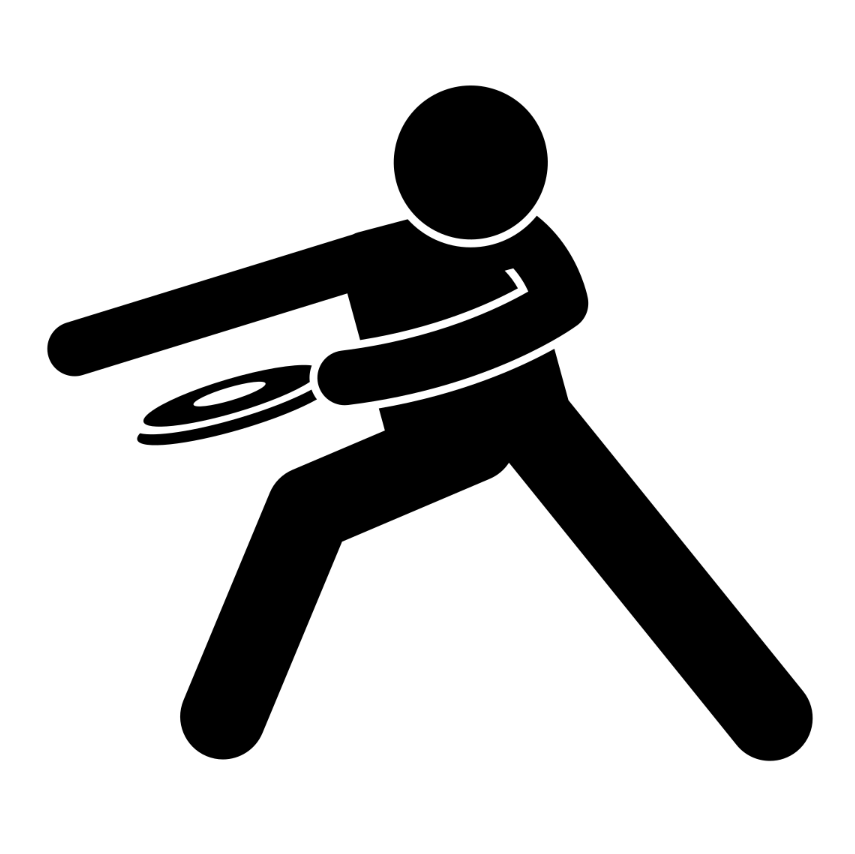}}      & \raisebox{-.5\height}{\includegraphics[width=1cm]{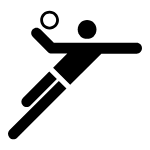}} &
           \raisebox{-.5\height}{\includegraphics[width=1cm]{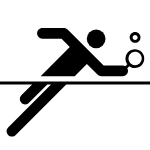}} & \raisebox{-.5\height}{\includegraphics[width=1cm]{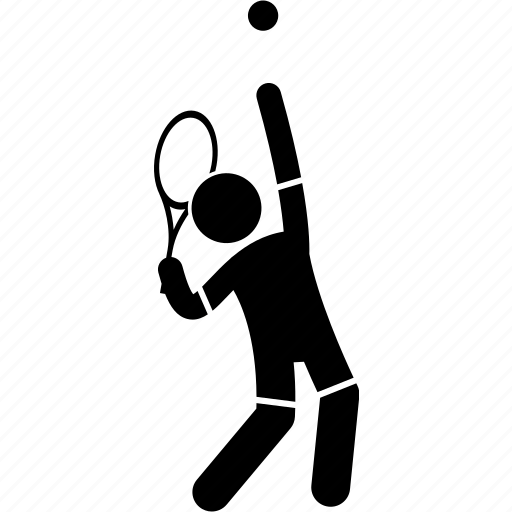}} & \raisebox{-.5\height}{\includegraphics[width=0.9cm]{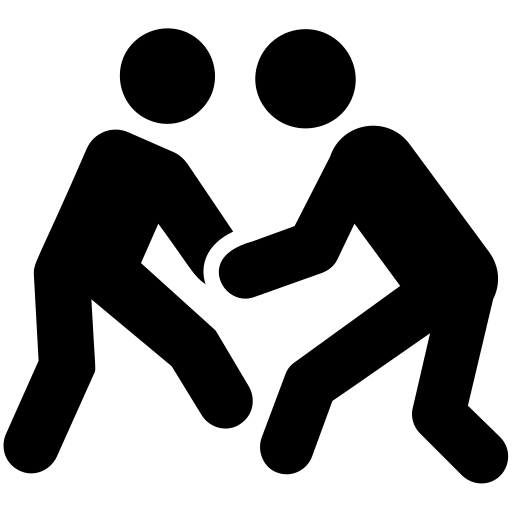}} \\ \hline 
\multirow{3}*{\textit{User Study}} &
MotionCLIP~\cite{tevet2022motionclip} & 8.00\% & 12.00\% & 6.00\% & 2.00\% & 10.00\% & 6.00\% & 8.00\% & 4.00\% & 4.00\% & 6.00\% & 16.00\%   \\  
&AvatarCLIP~\cite{hong2022avatarclip} & 10.00\% & 6.00\% & 10.00\% & 8.00\% & 6.00\% & 10.00\% & 12.00\% & 8.00\% & 2.00\% & 12.00\% & 18.00\%  \\  
&Ours  & \pmb{82.00\%} & \pmb{82.00\%} & \pmb{84.00\%} & \pmb{90.00\%} & \pmb{84.00\%} & \pmb{84.00\%} & \pmb{80.00\%} & \pmb{88.00\%} & \pmb{94.00\%} & \pmb{82.00\%} & \pmb{66.00\%} 
\\ \hline\hline
      Metrics & Methods&    \raisebox{-.5\height}{\includegraphics[width=1cm]{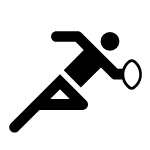}} & \raisebox{-.5\height}{\includegraphics[width=1cm]{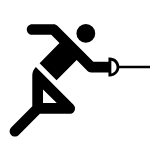}} & \raisebox{-.5\height}{\includegraphics[width=0.85cm]{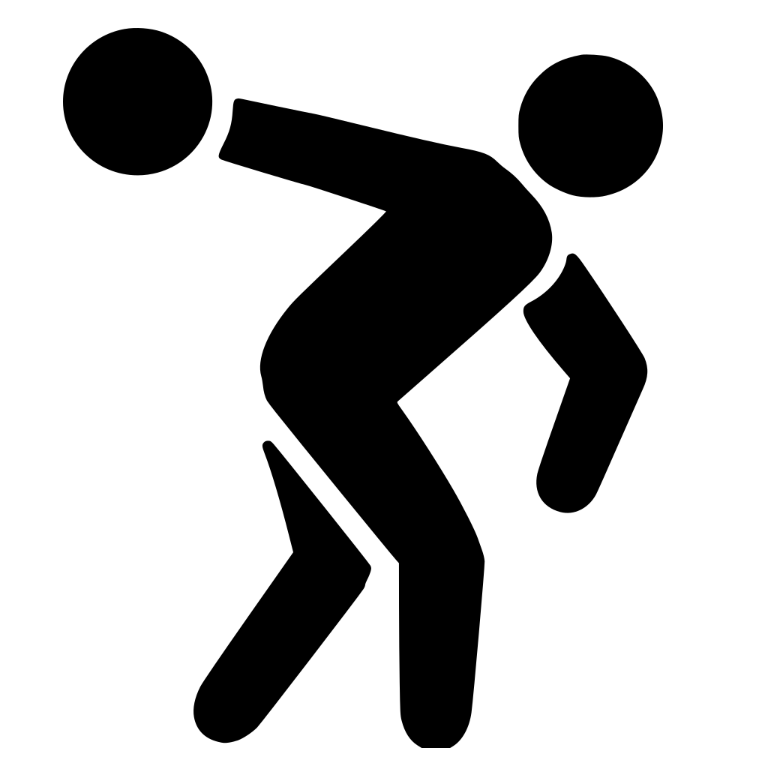}} & \raisebox{-.5\height}{\includegraphics[width=1cm]{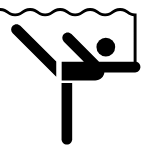}} & \raisebox{-.5\height}{\includegraphics[width=1cm]{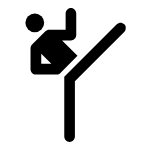}}     & \raisebox{-.5\height}{\includegraphics[width=1cm]{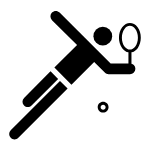}} & \raisebox{-.5\height}{\includegraphics[width=1cm]{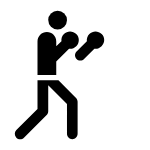}}       & \raisebox{-.5\height}{\includegraphics[width=1cm]{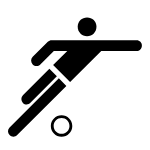}}   & \raisebox{-.5\height}{\includegraphics[width=1cm]{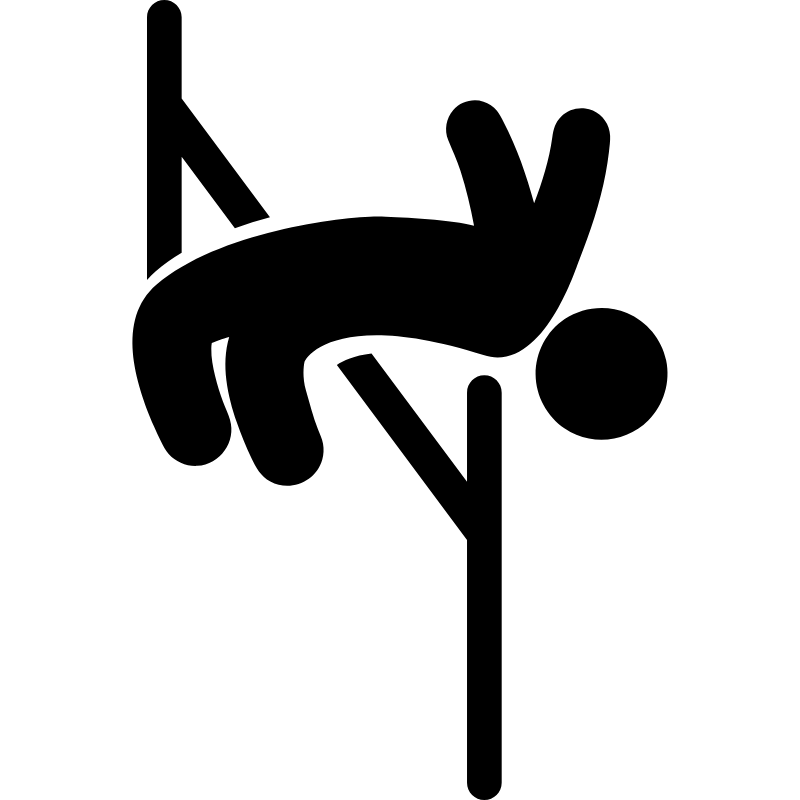}} & \raisebox{-.5\height}{\includegraphics[width=1cm]{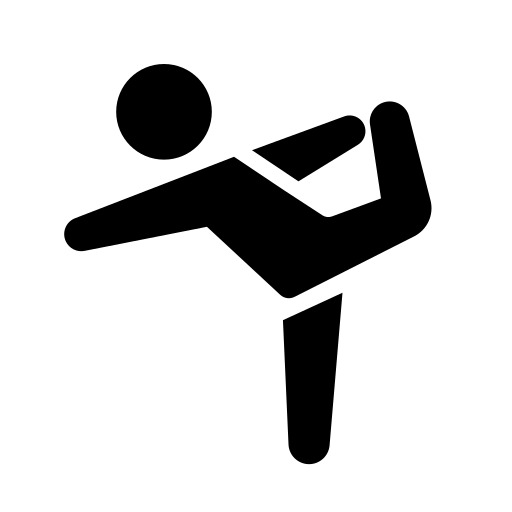}}  & \raisebox{-.5\height}{\includegraphics[width=1cm]{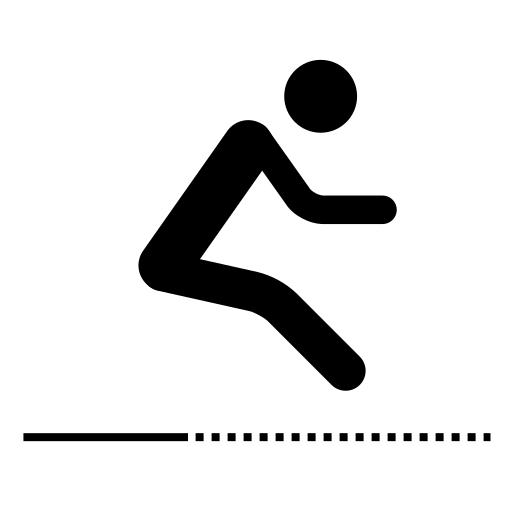}} \\ \hline  
\multirow{3}*{\textit{User Study}} &
MotionCLIP~\cite{tevet2022motionclip} & 6.00\% & 4.00\% & 4.00\% & 12.00\% & 8.00\% & 2.00\% & 14.00\% & 2.00\% & 26.00\% & 20.00\% & 14.00\%   \\  
&AvatarCLIP~\cite{hong2022avatarclip} & 4.00\% & 2.00\% & 8.00\% & 20.00\% & 6.00\% & 12.00\% & 6.00\% & 4.00\% & 34.00\% & 32.00\% & 18.00\%  \\  
&Ours  & \pmb{90.00\%} & \pmb{94.00\%} & \pmb{88.00\%} & \pmb{68.00\%} & \pmb{86.00\% }& \pmb{86.00\%} & \pmb{80.00\%} & \pmb{94.00\%} & \pmb{40.00\%} & \pmb{48.00\%} & \pmb{68.00\%} 
\\ \hline
\end{tabular}
}
\vspace{-5mm}
\end{table*}

\subsection{Style Transfer}
\begin{wraptable}{r}{6cm}
\vspace{-6mm}
\caption{\textbf{Style Transfer.} FreeMotion surpasses existing methods significantly.}
\vspace{-6mm}
\label{tab:style_transfer}
\begin{center}
\resizebox{\linewidth}{!}{
\begin{tabular}{lccc}
  \toprule
    & \multicolumn{3}{c}{\textit{User Study}}\\
    & MotionCLIP~\cite{tevet2022motionclip} & AvatarCLIP~\cite{hong2022avatarclip} & Ours\\ \midrule
  Happy & 22.67\% & 25.33\% & \pmb{52.00\%} \\
  Proud & 24.00\% & 18.00\% & \pmb{58.00\%}\\
  Angry & 14.00\% & 34.67\% & \pmb{51.33\%}\\
  Childlike & 28.67\% & 29.33\% & \pmb{42.00\%}\\
  Depressed & 14.67\% & 17.33\% & \pmb{68.00\%}\\
  Drunk & 11.33\% & 9.33\% & \pmb{79.33\%}\\
  Old   & 17.33\% & 28.00\% & \pmb{54.67\%}\\
  Heavy & 20.00\% & 16.00\% & \pmb{64.00\%}\\ \midrule
  Average & 19.08\% & 22.25\% & \pmb{58.67\%}\\
  \bottomrule
\end{tabular}}
\vspace{-16mm}
\end{center}

\end{wraptable}
\vspace{-3mm}
We evaluate our method's ability to represent motion styles without any training data. For this evaluation, we closely adhere to the settings established in MotionCLIP~\cite{tevet2022motionclip}, generating actions with specific styles directly from textual descriptions. This evaluation encompasses three action categories: Jump, Walk, and Stand, each expressed in eight distinct styles. We adopt the user study as before.
\vspace{-2mm}
\subsubsection{Analysis}
\begin{figure*}
	\centering	
 	\includegraphics[width=0.85\linewidth]{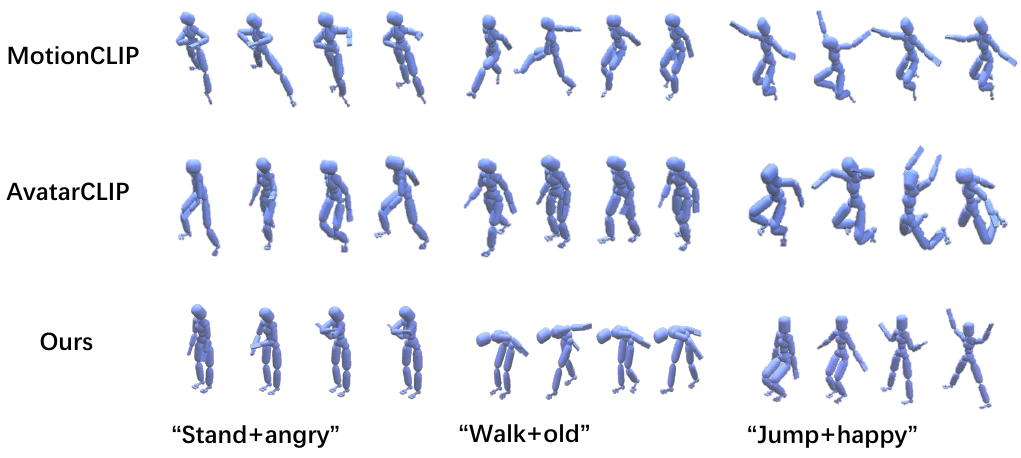}	
	\caption{\textbf{Visualization results of style transfer.} FreeMotion can add style to human motion using its world knowledge.
	}
	\label{fig:style_transfer}
\vspace{-5mm}
\end{figure*}

\vspace{-2mm}
The average user preference score of each style is shown in Tab.~\ref{tab:style_transfer}. FreeMotion won more than half of the votes. One impressive capability of MLLM is imagining what one will do in a specific style. For example, one will walk with a stoop when he is old. MotionCLIP and AvatarCLIP can hardly make it since they don't have explicit world knowledge and reasoning ability using natural language. The visualization results are shown in Fig.~\ref{fig:style_transfer}. Though sometimes CLIP-based baseline method can generate visually realistic frames such as the "Jump+happy" of AvatarCLIP in Fig.~\ref{fig:style_transfer}, the whole motion clip is not physically plausible and has low quality. 

\vspace{-3mm}
\subsection{Human-Scene Interaction}
Human-Scene Interaction presents a significant challenge in computer animation, necessitating not only the recognition of objects for interaction but also the generation of contextually appropriate motions. Specifically, for sitting and lying down, we utilize approximately 50 diverse items, including chairs, sofas, and beds, sourced from ShapeNet~\cite{chang2015shapenet}, and direct the humanoid to appropriately sit or lie on them. Similarly, for reaching tasks, we select around 50 different objects from ShapeNet and instruct the humanoid to reach them with a hand. The humanoid's initial position is set at a random distance from the object with a random orientation. We ran 40 times for each task.
\vspace{-2mm}
\subsubsection{Baseline}
\vspace{-2mm}
Data-driven methods~\cite{xiao2023unified,hassan2023synthesizing} solve Human-Scene Interaction synthesis by combining style reward from unlabeled motion data and task reward from human-designed reward function as in AMP~\cite{peng2021amp}. UniHSI~\cite{xiao2023unified} formulates the task reward in human-scene interaction tasks as Chain of Contacts. It obtains knowledge from LLMs to reason the contact pairs. For the different settings and the unavailability of code, we report the results from~\cite{xiao2023unified, hassan2023synthesizing} for a rough comparison. We also implement an AMP-based baseline trained on SAMP~\cite{hassan2021stochastic} for a fairer comparison. 

\begin{wraptable}{r}{7.5cm}
  \begin{center}
  \vspace{-18mm}
  \caption{\textbf{Human-Scene Interaction.} FreeMotion achieves good results on three interaction tasks.}
  \vspace{-3mm}
  \label{tab:hsi}
  \resizebox{\linewidth}{!}{
    \begin{tabular}{l|c|c|c|c|c|c}
    \toprule[1.5pt]
    \multirow{2}*{Methods}
    & \multicolumn{3}{c|}{Success Rate (\%) $\uparrow$}
    & \multicolumn{3}{c}{Contact Error $\downarrow$}\\
    & Sit & Lie Down & Reach & Sit & Lie Down & Reach \\
    \midrule
    InterPhys - Sit~\cite{hassan2023synthesizing}  & 93.7 & - & - & 0.09 & - & - \\
    InterPhys - Lie Down~\cite{hassan2023synthesizing} & - & 80.0 & - & - & 0.30 & - \\
    UniHSI~\cite{xiao2023unified} & 94.3 & \pmb{81.5} & \pmb{97.5} &\pmb{0.032} &\pmb{0.061}&0.016 \\
    \midrule
    AMP-Sit~\cite{peng2021amp}   & 83.6 & -    & -    & 0.074 & -     & -     \\
    AMP-Lie Down~\cite{peng2021amp}    & -    & 28.3    & - & - & 0.334     & -     \\
    AMP-Reach~\cite{peng2021amp}    & -    & -   & 96.6 & - & -    & 0.041     \\
    \midrule
    Ours                            & \pmb{95} & 60 & 95 & 0.066 & 0.224 & \pmb{0.012} \\
    \bottomrule[1.5pt]
    \end{tabular}}
  \vspace{-8mm}
  \end{center}
\end{wraptable}
\vspace{-2mm}
\subsubsection{Metric}
\vspace{-2mm}
In this task, we follow previous works~\cite{xiao2023unified, hassan2023synthesizing} that use \textit{Success Rate} and \textit{Contact Error} as the main metrics. However, these metrics should be computed with ground truth contact pairs, which are not available in our method. We manually set the contact pairs in the form of joints and target positions and inform GPT-4V in the prompt. 
\vspace{-2mm}
\subsubsection{Analysis}
\begin{figure*}
\vspace{-3mm}
    \centering
    \begin{minipage}{0.48\linewidth}
        \includegraphics[width=\linewidth]{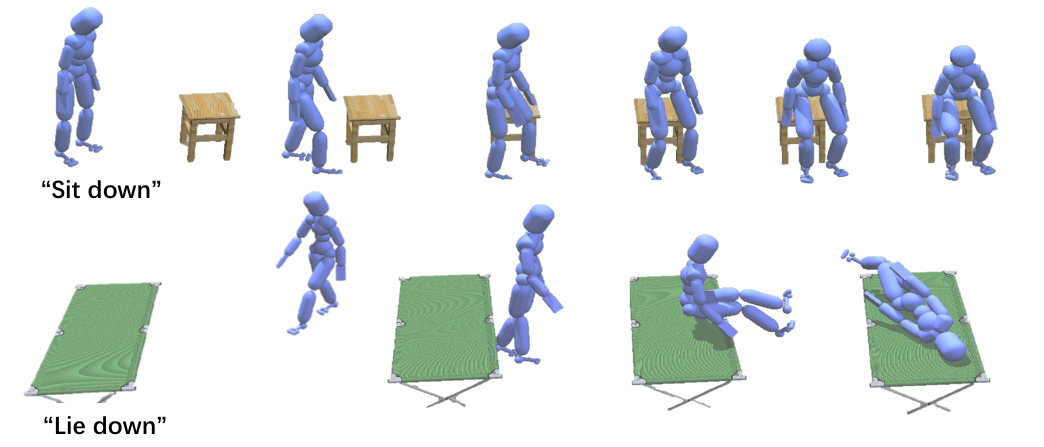}
        \caption{\textbf{Visualization of Human-scene interaction.} FreeMotion can navigate to and interact with the target object.}
        \label{fig:hsi}
    \end{minipage}\hfill
    \begin{minipage}{0.48\linewidth}
        \includegraphics[width=\linewidth]{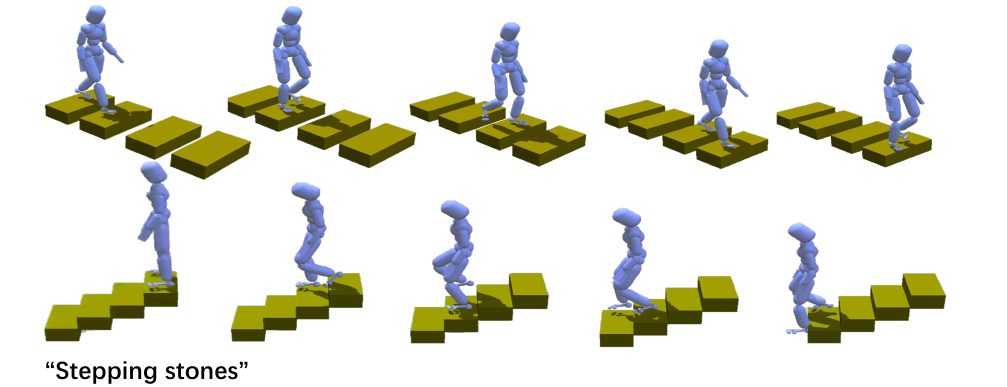}
        \caption{\textbf{Visualization results of stepping stones.} FreeMotion can navigate over irregular terrain.}
        \label{fig:stepping_stones}
    \end{minipage}
    \vspace{-5mm} 
\end{figure*}

\vspace{-2mm}
The results, as shown in Tab.~\ref{tab:hsi}, indicate that our method attains results comparable to previous methods, even in the absence of any motion data, which further implies that the MLLMs possess an innate understanding of scene interaction and object contact. However, the success rate decreases when lying down on a bed. This phenomenon can be mainly attributed to the rich contact in this process. Fig.~\ref{fig:hsi} illustrates the visualization results, where FreeMotion effectively guides the humanoid in navigating towards and interacting with the target object.
 \vspace{-5mm}
\subsection{Stepping Stones}
\vspace{-2mm}
Navigating challenging, irregular terrain is crucial for locomotion, with each footstep subject to strict constraints in this task. In this experiment, we adopt ALLSTEPS~\cite{xie2020allsteps} and select the best-performing policy (Adaptive) in the work for comparison.
\begin{wraptable}{r}{5cm}
\vspace{-12mm}
\caption{\textbf{Stepping Stones.} Please see the text for a detailed explanation of the numbers.}
\vspace{-6mm}
\label{tab:stepping_stones}
\begin{center}
\resizebox{\linewidth}{!}{
\begin{tabular}{lcc}
  \toprule
\textit{Task Parameter}    & ALLSTEPS~\cite{xie2020allsteps}  & Ours\\ \midrule
    \textit{Flat} ($\Theta = 0$) \\ 
    \hspace{6mm}$\Phi = 0$ & \pmb{1.45, 1.50}& 1.40, 1.45\\ 
    \hspace{6mm}$\Phi = 20$ & 1.35, 1.40&\pmb{1.40, 1.40}\\ \midrule
  \textit{Single-step} ($\Phi = 0$) & \\ 
    \hspace{6mm}$\Theta = 50$ &\pmb{0.80, 0.80} & 0.60, 0.75\\ 
    \hspace{6mm}$\Theta = -50$ &0.90, 0.95 & \pmb{1.00, 1.10}\\\midrule
  \textit{Continuous-step} ($\Phi = 0$) & \\ 
   \hspace{6mm}$\Theta = 50$ &\hspace{2.6mm}---, 0.65 & \pmb{0.50, 0.65}\\ 
    \hspace{6mm}$\Theta = -50$ &0.65, 0.70 & \pmb{0.75, 0.85}\\ \midrule
  \textit{Spiral} ($\Phi = 20$) & \\ 
   \hspace{6mm}$\Theta = 30$ &\pmb{0.80, 0.85} & 0.40, 0.80\\ 
    \hspace{6mm}$\Theta = -30$ &1.00, 1.10 & \pmb{1.10, 1.30}\\
  \bottomrule
\end{tabular}}
\vspace{-12mm}
\end{center}
\end{wraptable}
\vspace{-6mm}
\subsubsection{Baseline}
\vspace{-2mm}
 ALLSTEPS~\cite{xie2020allsteps} learns stepping-stone skills by utilizing deep reinforcement learning and curriculum learning. Though requiring no motion data, it needs carefully designed task-specific reward functions and training strategies to achieve good results.
\vspace{-2mm}
\subsubsection{Metric}
\vspace{-2mm}
We denote pitch $\Theta$, yaw $\Phi$, and distance $d$ as the parameters of each step relative to the previous step. We repeat each scenario five times and record two numbers following~\cite{xie2020allsteps}. The first number represents the maximum value of $d$ for which the policy succeeds for all five runs. The second number represents the maximum value of $d$ for which the policy succeeds in at least one of the runs. A larger number generally means a better capability to walk on difficult terrain.

\vspace{-2mm}
\subsubsection{Analysis}
\vspace{-2mm}
 The results are shown in Tab.~\ref{tab:stepping_stones}. FreeMotion not only achieves comparable or superior results without motion data or complex reward design expertise but also excels in generating actions that are both naturally harmonious and physically feasible. As showcased in Fig.~\ref{fig:stepping_stones}, it adeptly navigates irregular terrain, producing movements that are in line with the physical constraints of the stepping stones, further emphasizing its realistic motion synthesis capabilities.

 \vspace{-3mm}
\section{Ablation Study}
  \vspace{-3mm}
In this section, we conduct ablation experiments on the 12 action categories from HumanAct12~\cite{guo2020action2motion} to evaluate the effectiveness of our main designs. The dataset and the metric were introduced in the Motion Synthesis Task.
 \vspace{-1mm}
\subsection{Keyframe Designer} 
  \vspace{-1mm}
The generation of keyframe descriptions is the core of our keyframe designer. In FreeMotion, we ask the MLLM to output a general full-body description with detailed body-part descriptions to decompose the keyframe spatially. In this experiment, we remove the detailed body-part descriptions, only outputting a general sentence of the full body, to evaluate the effectiveness of explicit spatial decomposition. The result is shown in Tab.~\ref{tab:body_part_desc}. Detailed body-part descriptions help the keyframe generation process in motion synthesis.
 \vspace{-1mm}
\subsection{Keyframe Animator} 
 \vspace{-1mm}
 In this part, we remove the visual feedback mechanism in our keyframe animator, without which the MLLM can only call the command once for each body part. The result is shown in Tab.~\ref{tab:visual_feedback}. Visual feedback allows the MLLM to further adjust the humanoid's pose and improve the motion quality.\\
\begin{minipage}{\textwidth}
\vspace{3mm}
 \begin{minipage}[t]{0.50\textwidth}
  \centering
     \makeatletter\def\@captype{table}\makeatother\caption{\textbf{Ablation on body-part desc.}}
      \vspace{-1mm}
\scalebox{0.85}{ 
\begin{tabular}{lc}
  \toprule
& \textit{User Study} \\ \midrule
w/o body-part desc. & 26.00\%\\
Ours & 74.00\%\\
  \bottomrule
\label{tab:body_part_desc}
\end{tabular}}
  \end{minipage}
  \begin{minipage}[t]{0.50\textwidth}
   \centering
  \makeatletter\def\@captype{table}\makeatother\caption{\textbf{Ablation on visual feedback.}}
  \vspace{-1mm}
\scalebox{0.85}{ 
\begin{tabular}{lc}
  \toprule
& \textit{User Study} \\ \midrule
w/o visual feedback & 32.00\%\\
Ours & 68.00\%\\
  \bottomrule
  \label{tab:visual_feedback}
\end{tabular}}
   \end{minipage}
\end{minipage}

 \vspace{-5mm}
\section{Conclusion}
 \vspace{-3mm}
In this work, we for the first time, without any motion data, explore open-set human motion synthesis using natural language instructions as user control signals based on MLLMs across any motion task and environment. Our method can potentially serve as an alternative to motion capture for collecting human motion data, especially when the cost of motion capture is huge (e.g., collecting human interaction with different scenes). 

Though we have evaluated the effectiveness of our method on many downstream tasks, its application can be expanded to more scenarios (e.g., human-human interactions, contact-rich human-object interaction). 

There is much progress to be made in investigating technologies to improve the performance of our framework. Currently, our method can not handle complex human motions (e.g., dancing) or long text instructions. Its performance will also downgrade when the contact is rich. Future researchers may consider finetuning an MLLM with expert human motion knowledge. More powerful pose adjustment technologies, sometimes even a neural network, can be utilized for the mapping between natural language description and human pose. We hope our work can pave the way for future work in this area.

\clearpage  

%
%
\bibliographystyle{splncs04}
\bibliography{main}

\title{FreeMotion: MoCap-Free Human Motion Synthesis with Multimodal Large Language Models} 

\titlerunning{Abbreviated paper title}

\author{First Author\inst{1}\orcidlink{0000-1111-2222-3333} \and
Second Author\inst{2,3}\orcidlink{1111-2222-3333-4444} \and
Third Author\inst{3}\orcidlink{2222--3333-4444-5555}}

\authorrunning{F.~Author et al.}

\institute{Princeton University, Princeton NJ 08544, USA \and
Springer Heidelberg, Tiergartenstr.~17, 69121 Heidelberg, Germany
\email{lncs@springer.com}\\
\url{http://www.springer.com/gp/computer-science/lncs} \and
ABC Institute, Rupert-Karls-University Heidelberg, Heidelberg, Germany\\
\email{\{abc,lncs\}@uni-heidelberg.de}}



\clearpage
\setcounter{page}{1}
\section*{\large{Supplementary Materials}}
\appendix

In the supplementary materials, we will first show more details and an example of our interaction with the MLLM in Sec.~\ref{supp:prompt}, including our keyframe designer and keyframe animator. Then implementation details, including our network architecture and the implementation of our command set, are provided in Sec.~\ref{supp:details}. We report detailed experiment settings in Sec.~\ref{supp:exp}, including user study and how the metrics are computed in human-scene interaction. We finally provide more analysis of FreeMotion in Sec.~\ref{supp:analysis}, including time consumption, motion diversity, and limitation. 

\section{Interaction with the MLLM}
\label{supp:prompt}
In this section, we provide more details and an example of the input and output of the MLLM. We use almost the same prompt for all of our downstream tasks, with minor differences to claim the definition of each task.
\subsection{Keyframe Designer}
Our keyframe designer decomposes a motion and generates a sequence of keyframes. We show the input and output of our keyframe designer in~\cref{tab:designer_input} and~\cref{tab:designer_output} respectively.

\begin{table*}
  \begin{center}
  \caption{\textbf{Example of keyframe designer's input.}}
  \label{tab:designer_input} 
  \resizebox{\textwidth}{!}{
    \begin{tabular}{p{420pt}}
    \toprule[1.5pt]
    Input \\
    \midrule
\begin{wrapfigure}{r}{2.3cm} 
\vspace{-9mm}
\includegraphics[width=2.3cm]{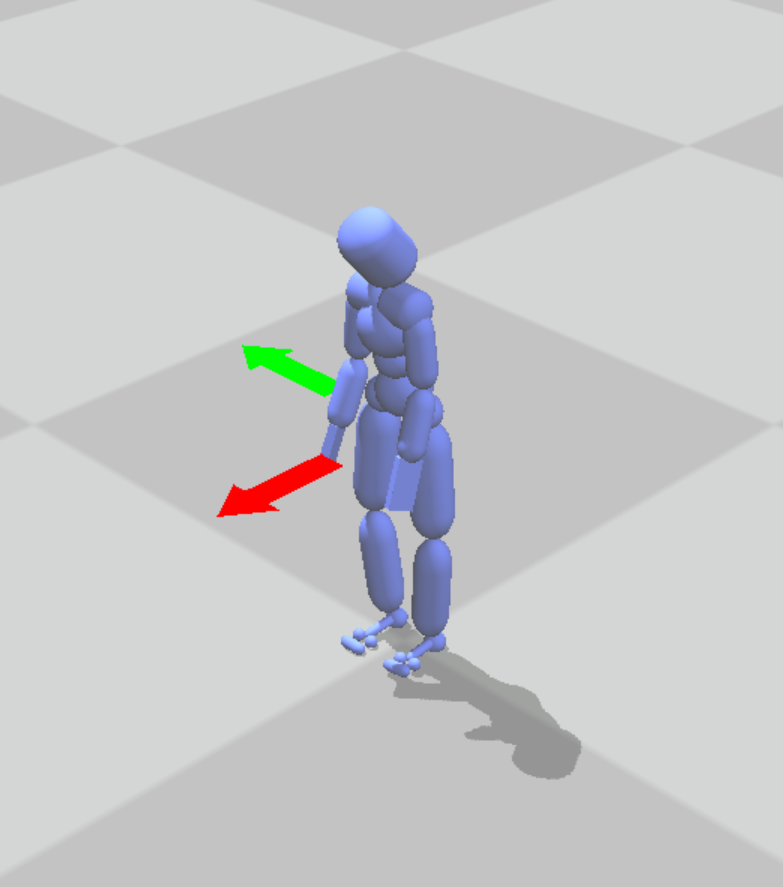}\
\end{wrapfigure}
I want you to be a human motion animation designer and design the keyframes of a humanoid to complete a motion. This is the picture of this humanoid. You are in the front left of him. A keyframe is a drawing or shot that defines the starting and ending points of any smooth transition. A motion can be represented by a series of keyframes. At each time step, you need to give a description of the next keyframe posture of the humanoid based on its current keyframe posture and picture. The following content should be concluded in your description: \\
1. General next-keyframe description: [DESCRIPTION: A general description of the next keyframe posture of the humanoid.]\\
2. [optional] Pelvis next-keyframe rotation: The pelvis of the humanoid rotates [NUM: 0.0] degrees to the [DIRECTION: left, right]. \\
3. [optional] Pelvis next-keyframe movement: The pelvis of the humanoid moves [NUM: 0.0] meters [DIRECTION: upward, downward], [NUM: 0.0] meters [DIRECTION: forward, backward], [NUM: 0.0] meters to the [DIRECTION: left, right]. \\
4. Left leg next-keyframe posture: [DESCRIPTION: A detailed description of the next keyframe posture of the humanoid’s left leg, including left knee, left ankle, and left toes.] \\
5. Right leg next-keyframe posture: [DESCRIPTION: A detailed description of the next keyframe posture of the humanoid’s right leg, including right knee, right ankle, and right toes.] \\
6. Torso next-keyframe posture: [DESCRIPTION: A detailed description of the next keyframe posture of the humanoid’s torso.] \\
7. Head next-keyframe posture: [DESCRIPTION: A detailed description of the next keyframe posture of the humanoid’s head.] \\
8. Left arm next-keyframe posture: [DESCRIPTION: A detailed description of the next keyframe posture of the humanoid’s left arm, including left shoulder, left elbow, left wrist, and left fingers.] \\
9. Right arm next-keyframe posture: [DESCRIPTION: A detailed description of the next keyframe posture of the humanoid’s right arm, including right shoulder, right elbow, right wrist, and right fingers.] \\
10. The time interval between the current state and the predicted next keyframe is [NUM: 0.0] seconds. \\
Rules: \\
1. The predicted keyframe posture should be achieved from the current posture by a smooth transition. \\
2. If a bullet point is marked [optional], do NOT add it unless it’s absolutely necessary \\
3. If you see phrases like [DESCRIPTION: requirement], replace the entire phrase with a descriptive sentence following the requirement. \\
4. If you see phrases like [NUM: default\_value], replace the entire phrase with a numerical value. \\
5. If you see phrases like [DIRECTION: direction1, direction2, ...], replace the entire phrase with a direction listed. \\
6. I will tell you the motion the humanoid needs to do. You MUST: 1) analyze the motion and information provided by the rendered picture; 2) output the analysis process before you make the description. When you are ready for the description, begin it with [start of description] and end it with [end of description]. \\
7. When you make the description of a body part, the following content should be included in your description: 1) where the body part moves and to what position; 2) the body part's relative positioning with other body parts, nearby environment, or its current posture; 3) the degree of curvature of the body part. \\
8. Be as detailed as possible. In each description, you should provide as much information as you can to describe the orientation and position of the corresponding body part. \\
9. Pelvis is the root joint of the humanoid. Its rotation and movement represent the overall rotation and movement of the humanoid. \\
10. When you think the motion has been completed or a periodic motion has concluded after a full cycle, replace the whole description with a single word "Done". \\
Information about the coordinate system: \\
1. The positive x-axis direction is pointed by the green arrow. \\
2. The up direction is the positive y-axis direction. \\
3. The positive z-axis direction is pointed by the red arrow. \\
4. The humanoid joint’s coordinate is a 3d vector [x, y, z]. \\
The motion needs to do is: Walk. The current posture of the humanoid is: the humanoid is standing on the ground. The humanoid’s current joint list and coordinates are \{We omit the coordinates here\}. The picture is shown here. Please help me with the task. Think step by step. \\

    \bottomrule[1.5pt]
    \end{tabular}}
  \end{center}
\end{table*}

\begin{table*}
  \begin{center}
  \vspace{-10mm}
  \caption{\textbf{Example of the keyframe designer's output.}}
  \label{tab:designer_output} 
  \resizebox{\textwidth}{!}{
    \begin{tabular}{p{420pt}}
    \toprule[1.5pt]
    Output \\
    \midrule
To animate a humanoid character walking, we need to analyze the mechanics of a walking motion, which typically involves a sequence of poses or "keyframes" that simulate the human gait. Walking is a cyclical activity where each cycle (step) is composed of two main phases for each leg: the 'stance phase,' where the foot is in contact with the ground, and the 'swing phase,' where the foot moves forward to take the next step. The humanoid's arms counterbalance the legs, moving in opposition to them. For simplicity, we'll assume a regular walking pace and begin with the humanoid's right leg and left arm forward and vice versa for the left leg and right arm.

Considering that the humanoid's current posture is standing, the first motion in a walk cycle for the next keyframe would involve shifting weight to one leg while moving the other leg forward. The corresponding arm opposite the forward leg will also move forward. The pelvis will tilt and rotate slightly to balance the motion, and the head will remain level with the horizon, eyes looking forward.

[Start of description]

General next-keyframe description: The humanoid's right leg moves forward, initiating the swing phase, while the left leg takes the weight, entering the stance phase. The right arm swings back while the left arm swings forward to maintain balance.

Pelvis next-keyframe movement: The pelvis of the humanoid moves 0.1 meters forward.

Left leg next-keyframe posture: The left knee is only slightly bent as it bears the weight of the body, the ankle is flexed to keep the foot flat on the ground, and the toes point straight ahead.

Right leg next-keyframe posture: The right knee lifts in preparation for the forward step, the ankle extends with the heel leading, and the toes point up and away from the ground.

Torso next-keyframe posture: The torso leans slightly forward with a straight back to maintain momentum and balance as the legs move.

Head next-keyframe posture: The head is upright, eyes looking forward, maintaining a level position as the body moves.

Left arm next-keyframe posture: The left shoulder is slightly raised, the elbow is bent at a natural angle, and the hand is in a relaxed position, moving forward and slightly up.

Right arm next-keyframe posture: The right shoulder drops slightly, the elbow straightens, and the hand moves backward and down in a mirror image of the left arm's movement.

The time interval between the current state and the predicted next keyframe is 0.5 seconds.

[End of description]

This description assumes the humanoid starts the walking motion with the right leg moving into the swing phase and the left leg staying in the stance phase. The arms move in opposition to their respective legs to maintain balance. The pelvis rotation and forward movement are subtle, as the pelvis anchors the upper body and leads the directional movement. The time interval of 0.5 seconds is a typical duration for one step in a casual walking pace.
 \\
    \bottomrule[1.5pt]
    \end{tabular}}
  \end{center}
\end{table*}

\subsection{Keyframe Animator}
Given a description from the keyframe designer, the keyframe animator calls a set of commands to adjust the humanoid's pose to fit the description. The adjustment is conducted in the order of different body parts listed by the keyframe designer, which is "pelvis, left leg, right leg, torso, head, left arm, right arm". Since the callable commands and available joints are not the same for each body part, (e.g., When adjusting the pelvis, the animator shouldn't call an end effector movement on the right ankle.) we utilize different sub-keyframe-animators for different body parts, varying in their access to the commands and joints. The correspondence between body parts and their available commands are shown in~\cref{tab:correspondence}. We also show an example of the right leg sub-keyframe-animator in ~\cref{tab:animator_input_output}.

\begin{table}
\vspace{-5mm}
\caption{\textbf{Correspondence between body parts and available commands.}}
\vspace{-6mm}
\label{tab:correspondence}
\begin{center}
\resizebox{\textwidth}{!}{
\begin{tabular}{l|l}
  \toprule
    body part & available commands\\ \midrule
  pelvis  & pelvis rotation/movement with/without support points on the ground, camera rotation   \\ \midrule
  left leg/right leg  & single joint movement, end effector movement, single joint roll, camera rotation \\ \midrule
  torso & single joint movement, single joint roll, camera rotation \\ \midrule 
  head  & single joint movement, single joint roll, camera rotation \\ \midrule
  left arm/right arm & single joint movement, end effector movement, single joint roll, camera rotation\\
  \bottomrule
\end{tabular}}
\end{center}
\vspace{-10mm}
\end{table}

\begin{table*}
  \begin{center}
  \caption{\textbf{Example of right leg sub-keyframe-animator's input and output.} Note that the adjustments of the pelvis and left leg have been finished. }
  \label{tab:animator_input_output} 
  \resizebox{\textwidth}{!}{
    \begin{tabular}{p{420pt}}
    \toprule[1.5pt]
    Input \\
    \midrule
\begin{wrapfigure}{r}{2.3cm} 
\vspace{-9mm}
\includegraphics[width=2.3cm]{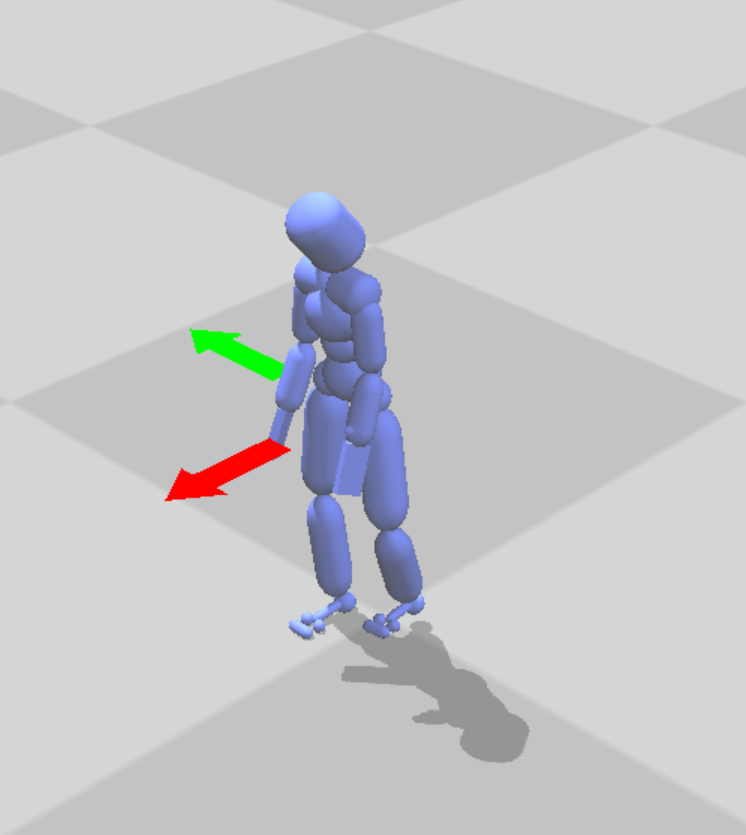}\
\end{wrapfigure}
I want you to be a humanoid motion animator. I will provide you with the humanoid’s current joint coordinates, a rendered picture, and a description of the target pose. You need to adjust the humanoid’s right leg pose using a set of commands to make it fit the description. This is the picture of this humanoid. You are in the front left of him. \\
Information about the coordinate system: \\
1. The positive x-axis direction is pointed by the green arrow. \\
2. The up direction is the positive y-axis direction. \\
3. The positive z-axis direction is pointed by the red arrow. \\
4. The humanoid joint’s coordinate is a 3d vector [x, y, z]. \\
Information about commands and how to use them: \\
1. Single joint movement. When you use this command, the selected joint will rotate around its parent joint to reach the target position. This is a basic command. You can use it if you want to change a certain joint’s position. Command format: command1, selected joint \{CHOICE: right\_knee, right\_ankle, right\_toes\}, movement \{[dx, dy, dz]: [0, 0, 0]\} \\
2. End effector movement. When you use this command, the selected end effector will try its best to reach the target position. You can use it if you want a certain end effector to move to a target position quicker and more directly. Command format: command2, selected end effector \{CHOICE: right\_toes\}, movement \{[dx, dy, dz]: [0, 0, 0]\} \\
3. Single joint roll. When you use this command, the selected joint will roll clockwise. You can use it if you want to roll a certain joint. Command format: command3, selected joint \{CHOICE: right\_hip, right\_knee, right\_ankle, right\_toes\}, roll\_degree \{$\alpha$: 0\} \\
4. Camera rotation. When you use this command, the camera will rotate clockwise from a top-down perspective. You can use it if you want to rotate the camera to get a better view of a body part. Command format: command4, rotate\_degree \{$\alpha$: 0\} \\
Rules: \\
1. A target pose description includes a general description of the whole body and a partial description of the right leg. You only need to consider how to adjust the right leg to make it fit the description. \\
2. You can use one of the above commands at a time in the corresponding command format. Begin it with [start of command] and end it with [end of command]. \\
3. Think step by step. Analyze the description, the humanoid’s joint coordinates, and the information provided by the rendered picture before you use the command. \\
4. When you use one of the commands, the humanoid will do as you say. I will provide you with updated joint coordinates and the rendered picture as feedback. You can continue the adjustment based on this feedback. \\
5. If you see phrases like \{CHOICE: choice1, choice2, ...\} in a command format, it means you should replace the entire phrase with one of the choices listed. \\
6. If you see phrases like \{[dx, dy, dz]: default\_movement\} in a command format, it means you should replace the entire phrase with a numerical array to indicate the movement of the joint in three axes. \\
7. If you see phrases like \{$\alpha$: default\_rotation\} in a command format, it means you should replace the entire phrase with a degree number to indicate the rotation. \\
8. If you think the right leg's pose fits the target pose description, output a single word "Done". \\
The general target pose description is: The humanoid's right leg moves forward, initiating the swing phase, while the left leg takes the weight, entering the stance phase. The right arm swings back while the left arm swings forward to maintain balance. The partial description of his right leg is: The right knee lifts in preparation for the forward step, the ankle extends with the heel leading, and the toes point up and away from the ground. The humanoid’s current joint list and coordinates are \{We omit the coordinates here\}. Please help me with the task.\\
\midrule
Output \\
\midrule
\{We omit the analysis of the MLLM here.\} 

[start of command] 

command2, selected end effector right\_toes, movement [0, 0.1, 0.1] 

[end of command] 

\{We omit the conclusion of the MLLM here.\} \\
    \bottomrule[1.5pt]
    \end{tabular}}
  \end{center}
\end{table*}

\section{Implementation Details}
\label{supp:details}
\subsection{Network Architecture}
We use almost the same architecture as~\cite{yao2022controlvae}. The parameters of our encoder, decoder, and world model are shown in \cref{tab:hyper}. 

\begin{table*}\centering
 \tabcolsep = 3mm
  \caption{\textbf{Hyperparameters of FreeMotion.}}
    \begin{tabular}{|c|c|c|}
        \toprule
        \multirow{ 5}{*}{Encoder} & Hidden Layers & 2 \\
        & Hidden Units & 1024 \\
        & Activation & ELU \\
        & Batchsize & 512 \\
        & Learning Rate & $10^{-5}$ \\
        \midrule
        \multirow{ 6}{*}{Decoder} & Hidden Layers & 3 \\
        & Hidden Units & 512 \\
        & Activation & ELU \\
        & Batchsize & 512 \\
        & Number of Experts & 6 \\
        & Learning Rate & $10^{-5}$ \\
        \midrule
        \multirow{ 5}{*}{World Model} & Hidden Layers & 4 \\
        & Hidden Units & 512 \\
        & Activation & ELU \\
        & Batchsize & 512 \\
        & Learning Rate & $0.002$ \\
        \bottomrule
    \end{tabular}
    \vspace{5mm}
    \label{tab:hyper}
\end{table*}
\subsection{Command Set}

In this section, we will introduce the technologies behind our command set and how our method calls each command.
\subsubsection{Single Joint Movement} When our keyframe animator calls this command, it selects a joint $\boldsymbol{j_{i}}$ and outputs an offset value in three axes $[dx, dy, dz]$ for $\boldsymbol{j_{i}}$. The target location is $\boldsymbol{p_{i}} + [dx, dy, dz]$, where $\boldsymbol{p_{i}}$ is the original position of $\boldsymbol{j_{i}}$. We denote the parent joint of $\boldsymbol{j_{i}}$ as $\boldsymbol{j_{k}}$. We solve an inverse kinematic problem with an IK chain starting from $\boldsymbol{j_{k}}$ and ending in $\boldsymbol{j_{i}}$ using gradient descent to get $\boldsymbol{j_{i}}$ close to the target location.
\subsubsection{End Effector Movement} For faster adjustments of the humanoid pose, we implement a command to quickly move a selected end effector (toes and fingers) to a target place through a pre-defined IK chain. We define four IK chains for four end effectors. They start from "left shoulder", "right shoulder", "left hip", "right hip" and end in "left fingers", "right fingers", "left toes", "right toes" respectively. When an end effector $\boldsymbol{j_{e}}$ is chosen and an offset value $[dx, dy, dz]$ is given. We use gradient descent to solve an inverse kinematic problem on the corresponding IK chain to get $\boldsymbol{j_{e}}$ close to the target location.
\subsubsection{Pelvis Rotation/Movement with Support Points on the Ground} When a humanoid moves or rotates its body, there is usually one or more support points (e.g., left toes, right ankle) on the ground to prevent the full body from falling. The keyframe animator outputs support points and translation/rotation of the pelvis to call this command. This command first applies the rotation or translation on the pelvis. Then we solve an inverse kinematic problem to restore the support points to their original locations. Other joints move accordingly through forward kinematics.
\subsubsection{Pelvis Rotation/Movement without Support Points on the Ground} Sometimes the humanoid moves or rotates without support points on the ground (e.g., the humanoid jumps into the air). When the keyframe animator calls this command, a translation or rotation is directly applied to the pelvis joint and other joints move accordingly through forward kinematics.
\subsubsection{Single Joint Roll} When the keyframe animator calls this command and predicts the rotation angle $\omega$, a selected joint rolls $\omega$ degrees clockwise.
\subsubsection{Camera Rotation} Occlusions may prevent the keyframe animator from observing the body part of interest. Thus we allow the animator to rotate the camera. Concretely, the camera moves in a horizontal circle around the humanoid, pointing at the pelvis. The vertical distance between the horizontal circle and the pelvis remains unchanged during camera rotation.

\section{Experiment Settings}
\label{supp:exp}
\subsection{User Study}
\begin{figure*}
	\centering	
 	\includegraphics[width=\linewidth]{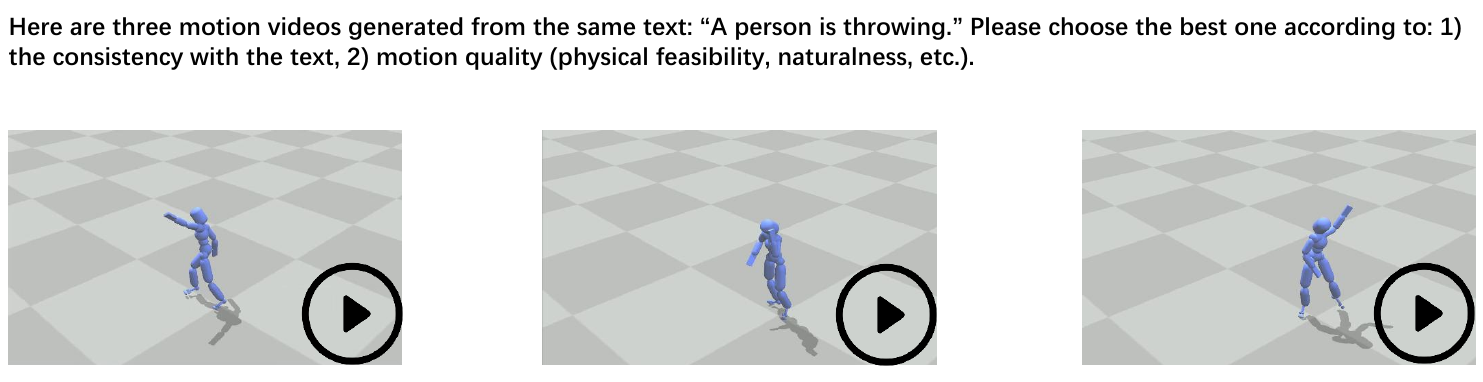}	
	\caption{\textbf{Example of user study interface.} We show the volunteer motions from FreeMotion and baseline methods side by side.}
	\label{fig:user_interface}
\end{figure*}

For motion synthesis and style transfer, we conduct user studies to evaluate our method. We generate four motions from baseline methods and FreeMotion respectively. For motion synthesis on HumanAct12, the prompt is an action category (e.g., "Walk") following the common use of HumanAct12. For motion synthesis on Olympic sports and style transfer, the prompt is a sentence starting with "A person". Then for baseline methods and FreeMotion, we randomly sample one from the four generated motions and show them to a college student volunteer side by side. The volunteer is asked to select the one with the best performance according to two focuses: 1) the consistency with input texts, and 2) motion quality (physical feasibility, naturalness, etc.). We collect 50 responses for each task. The example of our user study interface is shown in \cref{fig:user_interface}.
\subsubsection{Mapping of Olympic Sports}
The mapping between icons and sports names is shown in~\cref{tab:mapping}.
\begin{table*}[]
\caption{\textbf{Mapping between icons and sports names.}}
\vspace{-2mm}
\label{tab:mapping}
\resizebox{\textwidth}{!}{%
\begin{tabular}{ccccccccccc}
\hline
            \raisebox{-.5\height}{\includegraphics[width=1cm]{icons/skateboarding.png}} & \raisebox{-.5\height}{\includegraphics[width=1cm]{icons/lay_up.png}} & \raisebox{-.5\height}{\includegraphics[width=1cm]{icons/javalin_throw.png}} & \raisebox{-.5\height}{\includegraphics[width=1cm]{icons/jump_shot.png}}    & \raisebox{-.5\height}{\includegraphics[width=1cm]{icons/badminton.png}} & \raisebox{-.5\height}{\includegraphics[width=1cm]{icons/volley.png}} & \raisebox{-.5\height}{\includegraphics[width=0.95cm]{icons/frisbe.png}}      & \raisebox{-.5\height}{\includegraphics[width=1cm]{icons/handball.png}} &
           \raisebox{-.5\height}{\includegraphics[width=1cm]{icons/table_tennis.png}} & \raisebox{-.5\height}{\includegraphics[width=1cm]{icons/serve_tennis.png}} & \raisebox{-.5\height}{\includegraphics[width=0.9cm]{icons/wrestling.png}} \\ \hline 
Skateboarding & Lay up & Javalin throw & Jump shot & Badminton & Volley & Frisbe & Handball & Table tennis & Serve tennis & Wrestle
\\ \hline\hline
      \raisebox{-.5\height}{\includegraphics[width=1cm]{icons/rugby_sevens.png}} & \raisebox{-.5\height}{\includegraphics[width=1cm]{icons/fencing.png}} & \raisebox{-.5\height}{\includegraphics[width=0.85cm]{icons/discus_throw.png}} & \raisebox{-.5\height}{\includegraphics[width=1cm]{icons/rhythmic.png}} & \raisebox{-.5\height}{\includegraphics[width=1cm]{icons/taekwondo.png}}     & \raisebox{-.5\height}{\includegraphics[width=1cm]{icons/tennis.png}} & \raisebox{-.5\height}{\includegraphics[width=1cm]{icons/boxing.png}}       & \raisebox{-.5\height}{\includegraphics[width=1cm]{icons/football.png}}   & \raisebox{-.5\height}{\includegraphics[width=1cm]{icons/fosbury_flop.png}} & \raisebox{-.5\height}{\includegraphics[width=1cm]{icons/ballet.png}}  & \raisebox{-.5\height}{\includegraphics[width=1cm]{icons/long_jump.png}} \\ \hline  
Rugby sevens & Fencing & Discus throw & 
Gymnastics & Taekwondo & Tennis & Boxing & Football & High jump & Ballet & Long jump \\ \hline
\end{tabular}%
}
\vspace{-5mm}
\end{table*}

\subsection{Human Scene Interaction}
\textit{Success Rate} and \textit{Contact Error} are commonly used metrics in Human-Scene Interaction. A trial is detected as successful when a target joint is within 20 cm of the target location. \textit{Success Rate} is the percentage of successful trials. \textit{Contact Error} is the average distance between the target joint and the target location. Therefore, a contact pair, including a target joint and a target location, is needed for the computation of these metrics. We set the target joints as pelvis, head, and right fingers for sitting, lying down, and reaching respectively. A point on the target object's surface is manually sampled as the target location. We inform the GPT-4V of the contact pair in the prompt and ask it to obey this constraint when designing the motion. We ran 40 times on each task.
\section{Further Analysis of FreeMotion}
\label{supp:analysis}
\subsection{Time Consumption}
The time consumption of FreeMotion can be split into two parts, sequential keyframe generation from MLLMs and motion tracking. Typically, it takes dozens of inferences of the MLLM to generate a keyframe sequence. The time consumption at this stage depends heavily on the inference time of the MLLM, which is acceptable now and will continue to decrease in the future. It also takes about one hour to train a policy and corresponding world model to track one interpolated motion, which is a little bit time-consuming. However, a shared policy and world model, which are used in FreeMotion, can be trained by combining a batch of interpolated motions, which can largely reduce the time consumption per motion.
\subsection{Motion Diversity}
\begin{figure*}
	\centering	
 	\includegraphics[width=\linewidth]{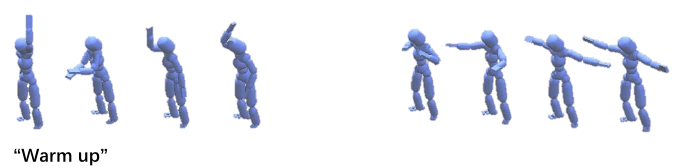}	
	\caption{\textbf{An example of motion diversity.} FreeMotion can synthesize diverse motions given the same input text.}
	\label{fig:warm_up}
\vspace{-5mm}
\end{figure*}

We also surprisingly find that diverse motions of the same category can be generated by FreeMotion thanks to the world knowledge of the MLLM. We show an example in \cref{fig:warm_up}. For warm-up action, different motions (extending arms horizontally or swinging arms vertically) can be generated using the same prompt input.
\subsection{Limitation}
Though we have evaluated the effectiveness of FreeMotion on many downstream tasks, its potential may not be fully exploited and the boundaries of its capability are not well-defined. Currently, we find FreeMotion struggles when dealing with long-text motion prompts (e.g., a person bounces on the balls of their feet and performs a couple of jabs with a closed fist and then practices protecting their side and head.) and complex motions (e.g., dancing). For long-text input, we find it hard for the MLLM to 1) determine the spacing between adjacent keyframes correctly and 2) fully understand the requirement of the prompt. For motions like dancing, the motions' complexity brings difficulties for keyframe generation and the generated full motions generally lack a sense of dynamics. We leave these problems to future works in human motion synthesis and more powerful MLLMs.   
%
%

\end{document}